\documentclass[journal]{IEEEtran}
\usepackage{graphics} 
\usepackage{amsmath} 
\usepackage{amssymb}  
\usepackage[colorlinks, linkcolor=blue, citecolor=blue]{hyperref}
\usepackage{cite}
\usepackage{graphicx}
\usepackage{color}
\usepackage{caption}
\usepackage{subfigure}
\usepackage{amsmath}
\usepackage{array}
\usepackage{multirow}
\usepackage{makecell}
\usepackage{xcolor}
\usepackage{mathtools}
\usepackage{amsmath,bm}
\usepackage{nomencl}
 \makenomenclature
\usepackage{booktabs} 
\usepackage{multirow}
\usepackage{mathrsfs}
\usepackage{array}

\ifCLASSINFOpdf
\else
\fi
\begin{document}
%
\title{LiDAR–Camera Calibration under Arbitrary Configurations: Observability and Methods}
%
%
%

\author{Bo~Fu,
	   Yue~Wang,~\IEEEmembership{Member,~IEEE,}
        Xiaqing~Ding,
		Yanmei~Jiao,
        Li~Tang,
        and~Rong~Xiong,~\IEEEmembership{Member,~IEEE}
\thanks{This work was supported in part by the National Key R\&D Program of China (2017YFB1300400), and in part by the National Nature Science Foundation of China (U1609210).}
\thanks{All authors are with the State Key Laboratory of Industrial Control and Technology, Zhejiang University, Hangzhou, P.R. China.
Yue Wang is the corresponding author {\tt\small wangyue@iipc.zju.edu.cn}.  Rong Xiong is the co-corresponding author {\tt\small rxiong@zju.edu.cn}. }
}


\maketitle

\begin{abstract}
LiDAR-camera calibration is a precondition for many heterogeneous systems that fuse data from LiDAR and camera. {\color{blue}{However, the constraint from  common field of view and the requirement for strict time synchronization make the calibration a challenging problem. In this paper, we propose a novel LiDAR-camera calibration method aiming to eliminate these two constraints. Specifically, we capture a scan of 3D LiDAR when both the environment and the sensors are stationary, then move the camera to reconstruct the 3D environment using the sequentially obtained images. Finally, we align 3D visual points to the laser scan based on tightly couple graph optimization method to calculate the extrinsic parameters between LiDAR and camera. Under this design, the configuration of these two sensors are free from the common field of view constraint owing to the extended view from the moving camera. And we also eliminate the requirement for strict time synchronization as we only use the single scan of laser data when the sensors are stationary. We theoretically derive the conditions of minimal observability for our method and prove that the accuracy of calibration is improved by collecting more observations from multiple scattered calibration targets. We validate our method on both simulation platform and real-world datasets. Experiments show that our method achieves higher accuracy than other comparable methods, which is  in accordance with our theoretical analysis. In addition, the proposed method is beneficial to not only plane measurement error based chessboard, but also other point measurement error based calibration targets, such as boxes and polygonal boards.}}
\end{abstract}


%
\IEEEpeerreviewmaketitle


\nomenclature{$\{\mathcal{C}\}$}{The camera coordinate system.}%

\nomenclature{$\{\mathcal{L}\}$}{The laser coordinate system.}%

\nomenclature{$^{\mathcal{C}}p_{c}$}{A visual 3D point on the chessboard in $\{\mathcal{C}\}$.}%

\nomenclature{$^{\mathcal{L}}p_{c}$}{A visual 3D point on the chessboard in $\{\mathcal{L}\}$.}%

\nomenclature{$^{\mathcal{L}}p_{f}$}{A visual 3D point of a landmark in $\{\mathcal{L}\}$.}%

\nomenclature{$p_{L}$}{{\color{blue}{A laser 3D point in $\{\mathcal{L}\}$.}}}%

\nomenclature{$p_{r}$}{A laser 3D point on the chessboard in $\{\mathcal{L}\}$.}%

\nomenclature{$n_{r}$}{The normal vector of a laser point.}%

\nomenclature{$^{\mathcal{L}}x_{c}$}{The pose of the camera in $\{\mathcal{L}\}$.}%

\nomenclature{$_{\mathcal{C}}^{\mathcal{L}}x$}{The extrinsic parameters of $\{\mathcal{C}\}$ w.r.t. $\{\mathcal{L}\}$.}

\printnomenclature

\section{Introduction}
%
%
%
%

\begin{figure}[tp]
\centering
      \includegraphics[scale=0.4]{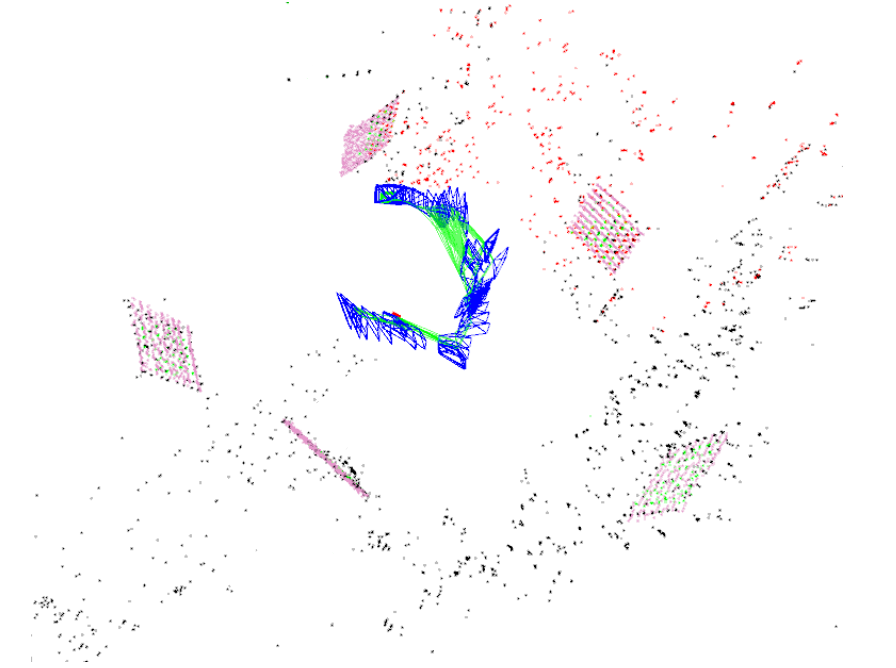}
      \caption{Projection of the LiDAR point cloud and visual point cloud using our calibration result. The green points are marked by the detected chessboard corners.}
	\label{placement}	
\end{figure}
\IEEEPARstart{A}{} perception system that employs only one sensor will not be robust. For example, LiDAR-based odometry \cite{zhuang20133} will fail when working in a long corridor, and the camera-based algorithm \cite{hussmann2009three} , \cite{zhu2010noncontact}, \cite{li2010measurement} cannot be applied to a textureless scene \cite{tang2018topological}. Fusing the visual and laser information can eliminate the outliers from the algorithm, and solve various limitations for the algorithms imposed by the single sensor. For example, the fusion of the range sensor and the camera can improve the accuracy of object detection \cite{luo20173d}. What's more, heterogeneous localization methods, such as visual localization on a laser map \cite{ding2018laser}, can enable low-cost and long-term localization. The precondition of all the above algorithms is the calibration of different sensors, and to that end, we focus on extrinsic calibration of the LiDAR and camera in this work.

Numerous efforts have been carried out to perform LiDAR–camera extrinsic calibration \cite{zheng2015new}, \cite{zhang2017real}, \cite{galilea2009calibration}. The current calibration approaches can be classified into two groups \cite{ishikawa2018lidar}: one is appearance-based and the other is motion-based. The appearance-based methods can obtain the extrinsic parameters by directly matching 2D images with 3D points on the laser point cloud. In the motion-based methods, the motion of the camera is estimated from images, while the motion of the LiDAR is estimated from the laser points, and then calibration is performed by aligning the two trajectories.

{\color{blue}{First, we will consider \textbf{the appearance-based methods}. Methods such as \cite{zhang2004extrinsic}, \cite{fremont2008extrinsic} use targets that can be detected on both 2D images and 3D laser point clouds. Geiger $et\ al.$ \cite{geiger2012automatic} presented a method to automatically calibrate the extrinsic parameters with one shot of multiple chessboards, which recovered the 3D structure from the detected image corners. After that, the approach used the constraint that the chessboard planes should coincide with the detected LiDAR planes to perform calibration. The method was applied in the KITTI dataset \cite{geiger2013vision} to calibrate the extrinsic parameters between the cameras and the LiDAR sensor. Unlike the approaches above, Wang $et\ al.$ \cite{wang2017reflectance} utilizes the reflectance intensity to estimate the corners of the chessboard from the 3D laser point cloud. If the corners of the 3D laser point cloud are identified, the extrinsic calibration is converted to a 3D-2D matching problem. However, these algorithms always require the sensors sharing a common field of view, which some application scenarios cannot satisfy. Even in the application scenario where the condition is met, the requirement of the common field of view constrains the scale of the scene and limits the number of targets that can be detected, thus affecting the accuracy of the calibration, which prevents the utilization of pinhole cameras from the LiDAR-camera system. In some methods, panoramic or wide-angle cameras are used to solve this problem \cite{mirzaei20123d}. Some methods lead to the tedious focus process in order to expand the field of view such as \cite{galilea2009calibration}.}}
\begin{figure*}[htbp]
     \centering
      \includegraphics[scale=0.5]{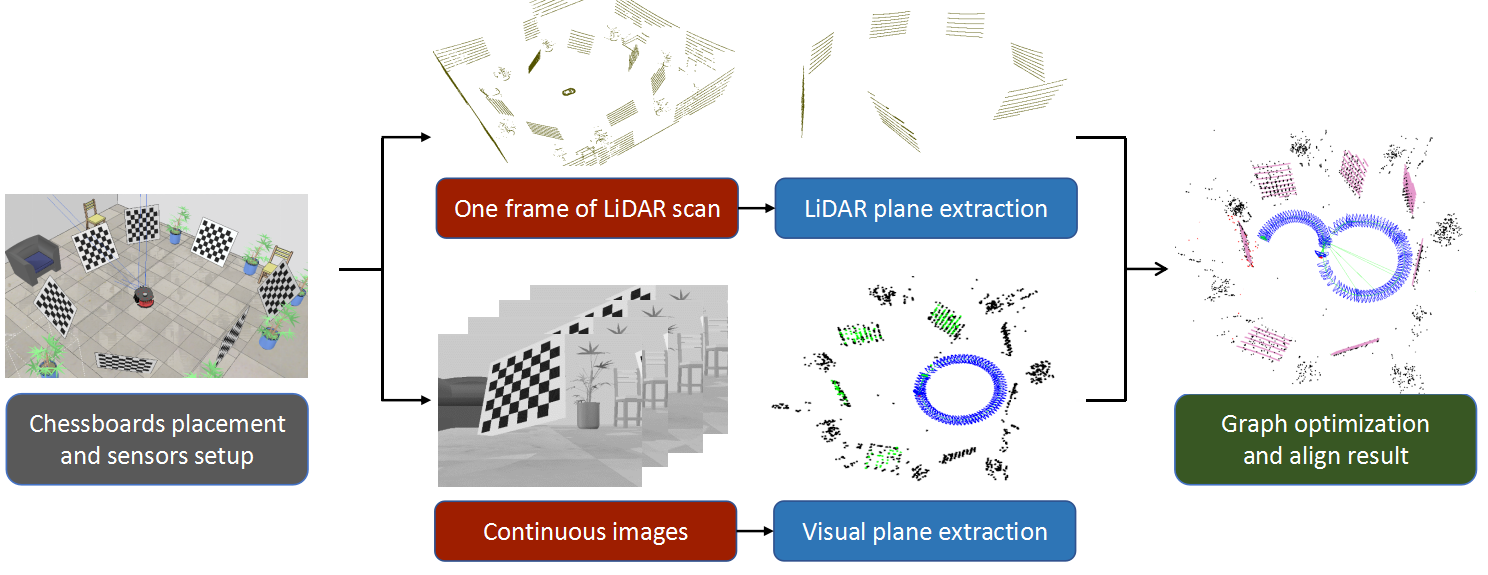}
      \caption{Overview (using a simulation experiment as an example). First, we fit the chessboard plane in the obtained laser data. Second, we extract the chessboard corner points from the camera images and then reconstruct the visual 3D point clouds. Third, we optimize the point-to-plane error to estimate the extrinsic parameters.}
      \label{overview}
\end{figure*}

{\color{blue}{On the other hand, \textbf{the motion-based methods} \cite{taylor2016motion}, \cite{shiu1989calibration} perform calibration by aligning the estimated motion trajectories. Early motion-based calibration methods were based on hand-eye calibration \cite{hol2010modeling}. In \cite{ishikawa2018lidar}, the initial extrinsic parameters are obtained from scale-free camera motion and LiDAR motion. Next, the camera motion is recalculated using the initial extrinsic parameters and the point cloud from the LiDAR, and then the extrinsic parameters are calculated again using the motion, and this is repeated until the estimate converges. However, the motion-based method is a loosely coupled calibration method that cannot lead to high calibration accuracy. In addition, the motion-based calibration method needs to complete time synchronization before performing calibration, which is not easy in some cases. In scenarios where time synchronization is not completed, an additional variable(i.e.\ time offset) should be introduced. In \cite{taylor2016motion}, they propose a method to obtain the motion of a sensor in 2D-3D calibration and estimate the extrinsic parameters and the time offset between the sensors. Obviously, introducing new variables will reduce the calibration accuracy.}}

In this work, we propose a hybrid calibration method, which combines the advantages of appearance-based calibration and motion-based calibration. The demonstration of proposed method is shown in Fig. \ref{placement}. In our method, a number of chessboards in various poses are placed around the sensors, and one frame laser scan of the chessboards is obtained under stationary. Then the sensors are moved around to obtain images of each chessboard to reconstruct the visual 3D point cloud. Note that this differs from previous approaches \cite{zheng2015new}, \cite{zhang2017real} which require multiple images and the LiDAR data of a single chessboard presented at different poses as inputs; the hidden limitation of these methods is that a common field of view between sensors is needed.

{\color{blue}{Our method expands the camera's field of view by moving the sensor, so even though there is no common view at the starting position, the LiDAR and the “expanded camera” can also have overlap in their measurement ranges, which removes the configuration limitation for a common field of view. Moreover, the extended field of view obtained can remove the constraints of the observed scale of the scene and increase the number of chessboards that can be detected, which can lead to an increase in accuracy. Additionally, since we only use the first frame of laser as a map. In this way, we eliminate the time variable (i.e. time offset) from the spatial extrinsic parameters estimating, which means that we don't need to solve time variable (i.e. time offset) and spatial variable (i.e. $_{\mathcal{C}}^{\mathcal{L}}x$) together. So our method is applicable to the cases lacking time synchronization and will not introduce additional variables. As part of our contribution, we also examine the observability properties of our system and present the minimal necessary conditions for concurrently estimating the LiDAR-camera extrinsic parameters. Further, we derive the influence of the angle and distance between calibration targets on the calibration accuracy, which proves that sharing a larger field of view between sensors is beneficial for better calibration accuracy. The relevant theory provides a guideline for designing high-accuracy calibration procedures.}}

This work is structured as follows. The next section starts with a discussion of the related work. Section II gives a detailed description of the proposed method. Then we prove the theory in Section III and Section IV and evaluate it in Section V. We present our conclusions in Section VI.

\section{Calibration Method}
{\color{blue}{Our optimization method is tightly coupled, and applicable to situations where there is no time synchronization.}} Additionally, in order to remove the configuration limitation, our method reconstructs visual point clouds from continuous images, which also expands the camera's field of view and improves calibration accuracy.

An overview of our method is shown in Fig. \ref{overview}. This method can be roughly divided into two steps. In the first step, we use the region growing method to segment the obtained laser data into several point cloud planes associated with the chessboard planes, and we use the corner extraction method to extract the chessboard corner points from the camera images and then reconstruct the visual 3D point clouds. Then we construct point-to-plane optimization equations from the visual points to the LiDAR plane, and estimate the extrinsic parameters using the Gauss–Newton method \cite{bjorck1996numerical}. In the following Section II-A, we will describe how both sets of points can be generated, and be aligned to get the extrinsic parameters in Section II-B.

\subsection{LiDAR And Visual Chessboard Plane Extraction}

Compared to existing methods for capturing multiple sets of images and LiDAR data for a single chessboard, we disperse multiple chessboards in space arranged in various poses.

{\color{blue}{Our method does not limit the calibration target. The reason we choose the chessboards as an example is that they are easy to obtain and cheap, which is same as \cite{song2008use}, \cite{liu2014robust}. In addition, our method can also use polygonal planar boards, boxes, etc., as the calibration targets. We completed the observability analysis and experimental comparison of different calibration targets as a complement to our work.}}

\subsubsection{LiDAR Plane Extraction}
This step describes how to extract the chessboard plane from one frame scan of the LiDAR. We perform the segmentation by growing and clustering the points in the laser data to several point sets which potentially correspond to the chessboard planes. {\color{blue}{First, the normal vector $n_{L}$ is computed for each laser point $p_{L}$. }} Second, several seed points $p_{r_{s}}$ are randomly selected from the laser points. Then we grow each random seed point greedily into each region represented by a corresponding point set $\bm{P_{r_{s}}}$. The growing principle is that if a point is a neighbor of the seed and its normal vector is similar to the seed's normal, the point will be added to the corresponding set of the seed.

After using the region growing method, we get several hypotheses of point sets extracted from the laser data $\bm{H} = \{\bm{P_{{r_{s}}}}\}$. Next, we filter out a subset $\bm{H_{chess}}$ where each point set represents a chessboard plane, which means the planes which are either insufficiently planar or significantly smaller than a chessboard will be removed. {\color{blue}{After this part, the laser point belonging to the chessboard is represented as $p_{r}$.}}

\subsubsection{Visual Plane Extraction}
To extract the 3D points of the chessboard from continuous camera images, we use the following method to extract the chessboard pixels and reconstruct them into 3D points. First, we run ORB-SLAM \cite{mur2017orb} using the acquired camera image to obtain the pose of the camera at each moment and the position of the 3D points in the map. In order to distinguish the 3D points of the chessboard, we extract the corners of the chessboard while processing each image. After detecting the corners of the chessboard, we mark the corresponding 3D points in the map. {\color{blue}{In the case of monocular camera, we use the scale of the chessboard, while in the case of stereo camera, the scale is known. We assume that the scale of the monocular camera is constant throughout the short experiment time interval.}} Because of the existence of pixel point observation error, we set a reconstruction score for each marked pixel point indicating the reconstruction quality:
\begin{equation}\label{eq.sub3}
  Score_{c}=N_{a}-{\color{blue}{\gamma}} \cdot N_{b}
\end{equation}
where $N_{a}$ is the number of keyframes that observe the point, and $N_{b}$ is the uncertainty of the depth of the point, {\color{blue}{and $\gamma$ is the scale factor chosen to keep the expected value of $N_{a}$ and $N_{b}$ approximately equal.}}

The pixel points marked with low scores will be discarded and are not part of the later process. {\color{blue}{After obtaining the filtered chessboard corner points, we perform bundle adjustment \cite{triggs1999bundle} to ensure that the resulting 3D points of the chessboard are relatively accurate in space.}} The visual 3D points obtained by this method can lead to a larger common field of view between the two sensors, which removes the configuration limitation and is shown to improve the calibration accuracy compared with the appearance-based calibration method in Section IV.

Our calibration system can be considered to be a visual localization on the laser map (i.e.\, the first frame of the laser). That is to say, the camera's pose for the first frame is ${^{\mathcal{L}}x_{c_{0}}}$, which is equal to $_{\mathcal{C}}^{\mathcal{L}}x$, so the global frame is the laser frame.
\subsection{Optimization For Calibration}
\begin{figure}[tbp]
      \includegraphics[width=0.35\textwidth]{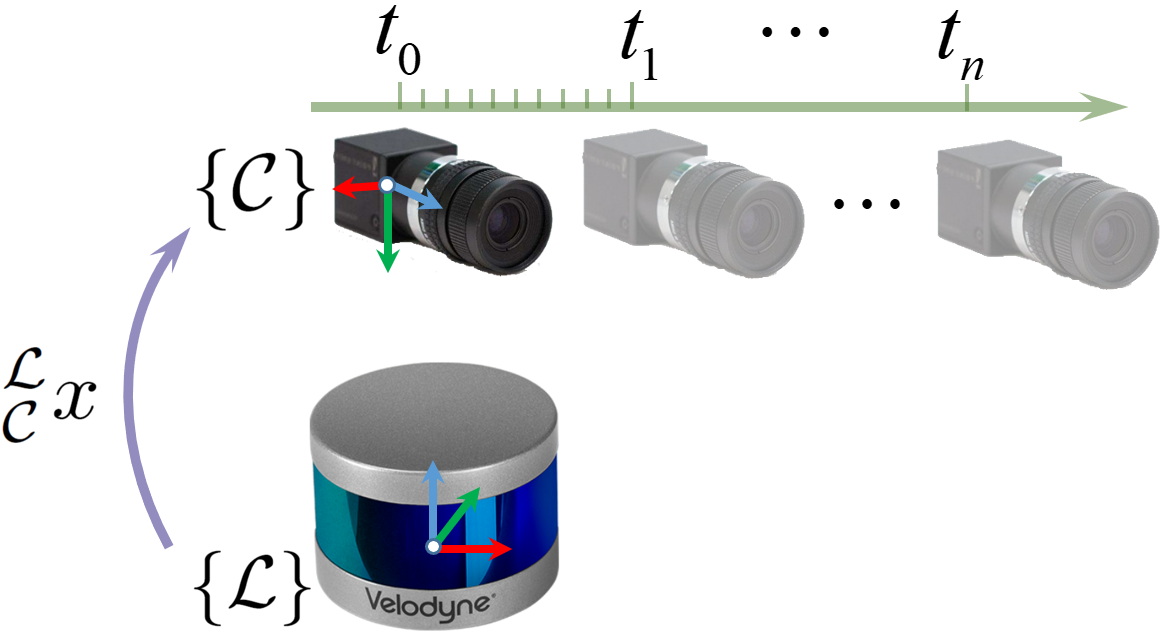}
      \caption{{\color{blue}{The definition of the coordinate system.}}}
      \label{hardware}
\end{figure}
The motion-based method performs calibration by aligning two trajectories which has a time synchronization problem. In order to explore the effects of time synchronization, we assume that the pose of LiDAR in $\{\mathcal{L}\}$ at time $t_{0}$ can be denoted by ${^{\mathcal{L}}x_{r_{0}}}$ , and ${^{\mathcal{L}}x_{r_{0}}}=I$. And the pose of the camera in $\{\mathcal{L}\}$ at time $t_{0}$ is ${^{\mathcal{L}}x_{c_{0}}}$, and ${^{\mathcal{L}}x_{c_{0}}}={_{\mathcal{C}}^{\mathcal{L}}x}$. In the ideal case, there would be an equation like this:
\begin{equation}\label{eq.subj1}
  {^{\mathcal{L}}x_{r_{0}}}={^{\mathcal{L}}x_{c_{0}}}\cdot{_{\mathcal{C}}^{\mathcal{L}}x}^{-1}
\end{equation}

Because of the existence of time synchronization error, the equation becomes:
\begin{equation}\label{eq.suj2}
   {^{\mathcal{L}}x_{\check{r_{0}}}}={^{\mathcal{L}}x_{c_{0}}}\cdot{_{\mathcal{C}}^{\mathcal{L}}x}^{-1}
\end{equation}
where ${^{\mathcal{L}}x_{\check{r_{0}}}}$ is the pose of LiDAR in $\{\mathcal{L}\}$ at time $t_{0}+\delta t$, and $\delta t$ is a small time offset. However, in our system, the two sensors remain stationary during laser data acquirement as shown in Fig. \ref{hardware}, so we can obtain:
\begin{equation}\label{eq.subj3}
  {^{\mathcal{L}}x_{r_{0}}}={^{\mathcal{L}}x_{\check{r_{0}}}}={^{\mathcal{L}}x_{c_{0}}}\cdot{_{\mathcal{C}}^{\mathcal{L}}x}^{-1}
\end{equation}

In this way, our calibration method is applicable for cases without time synchronization and will not introduce additional variables compared to the motion-based calibration method.

So far, 3D LiDAR chessboard points and 3D visual chessboard points have been obtained from the LiDAR data and continuous images, respectively. Next, we need to optimize the extrinsic parameters through the correspondences between these point clouds. First of all, the data association needs to be performed. These point clouds obtained in Section II-A are not one-to-one relevant. In fact, there is no need for point-to-point correspondence in our method; what we need is just chessboard-to-chessboard correspondence. Because the point clouds of the chessboard are very sparse, it is simple to get chessboard-to-chessboard correspondence. {\color{blue}{To obtain the data association, the mechanical parameters are used as the initial value of $_{\mathcal{C}}^{\mathcal{L}}x$.}} Then we use a K-dimension tree structure (KD-tree) to search for the nearest 3 laser points (i.e.\ , $p_{r_{1}}$, $p_{r_{2}}$, and $p_{r_{3}}$) for each $^{\mathcal{L}}p_{c}$ and save the above data association.

After the data association we need to filter the point pairs before putting them into the optimization process. We use the laser information to delete the visual points that should not be involved in optimization. Our score function is based on the distance from the visual point to the associated laser plane.
\begin{equation}\label{eq.sub4}
  Score_{r}=\sum_{i=1}^3 n_{r_{i}}^{T}(^{\mathcal{L}}p_{c}-p_{r_{i}})
\end{equation}
where $n_{r_{1}}$, $n_{r_{2}}$, and $n_{r_{3}}$ are the normal vectors corresponding to $p_{r_{1}}$, $p_{r_{2}}$, and $p_{r_{3}}$. We remove $^{\mathcal{L}}p_{c}$ with scores less than a certain threshold.

\begin{figure}[tp]
     \centering
      \includegraphics[width=0.48\textwidth]{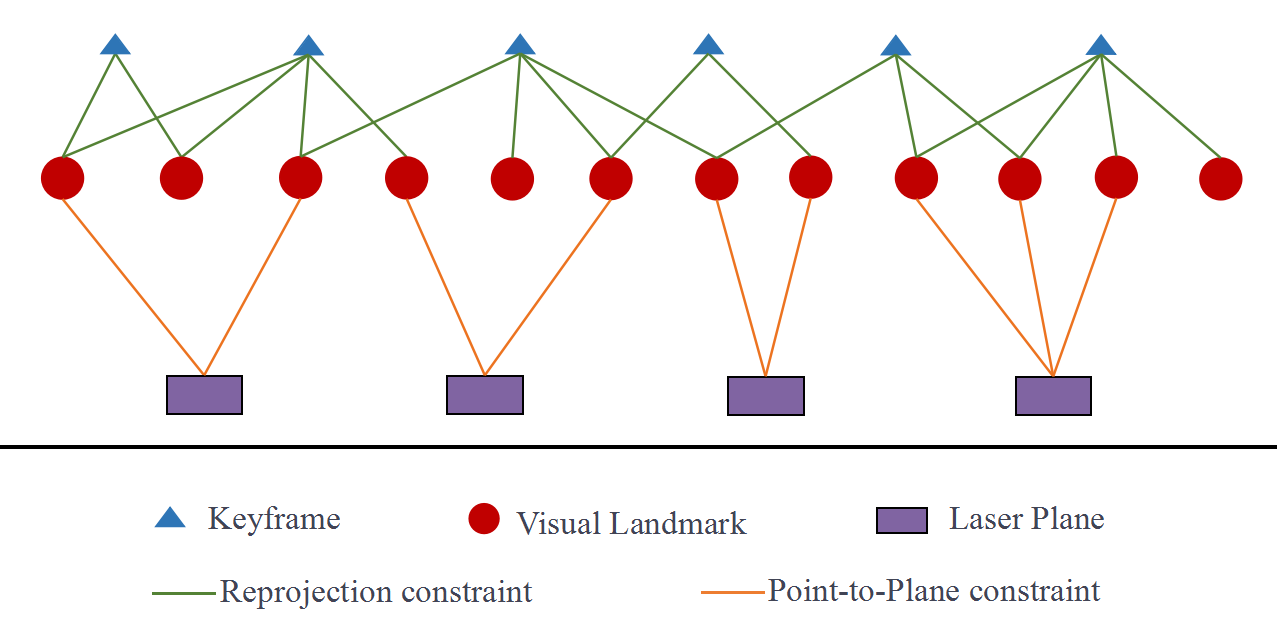}
      \caption{Optimization of the camera state $^{\mathcal{L}}x_{c}$, visual landmark $^{\mathcal{L}}p_{f}$ (the point belonging to the chessboard is represented as $^{\mathcal{L}}p_{c}$) in the global graph with the reprojection constraint and point-to-plane constraint. The laser point on the laser plane (i.e.\ , $p_{r}$) is an observation which will not change with time.}
      \label{optimize}
\end{figure}

Then we optimize the remaining points. As shown in Fig. \ref{optimize}, the state variables of the system are the camera state (i.e.\, the position of the keyframe) $^{\mathcal{L}}x_{c}$, and landmark $^{\mathcal{L}}p_{f}$ (the point belonging to the chessboard is represented as $^{\mathcal{L}}p_{c}$). $\bm{P_{c}}$ and $\bm{P_{f}}$ represent the point sets of $^{\mathcal{L}}p_{c}$ and $^{\mathcal{L}}p_{f}$, respectively, and their relations are $^{\mathcal{L}}p_{c}\in \bm{P_{c}} \subseteq \bm{P_{f}}$. {\color{blue}{The cost function used to optimize the state variables is derived as:}}
\begin{equation}\label{eq.sub5}
  E=\sum_{i,j} E_{proj}(^{\mathcal{L}}x_{c_{i}}, ^{\mathcal{L}}p_{f_{j}})+ \sum_{j} E_{pl}(^{\mathcal{L}}p_{c_{j}})
\end{equation}
where $E_{proj}(^{\mathcal{L}}x_{c_{i}}, ^{\mathcal{L}}p_{f_{j}})$ represents the feature reprojection error for the $i$-th camera pose and the $j$-th feature point.
\begin{equation}\label{eq.sub6}
\begin{split}
  E_{proj}(^{\mathcal{L}}x_{c_{i}}, ^{\mathcal{L}}p_{f_{j}})=\rho((\pi(^{\mathcal{L}}p_{f_{j}},^{\mathcal{L}}x_{c_{i}})-u_{i,j})^{T}\\
  \Omega_{i,j}(\pi(^{\mathcal{L}}p_{f_{j}},^{\mathcal{L}}x_{c_{i}})-u_{i,j})))
  \end{split}
\end{equation}
where $\rho(\cdot)$ is the Huber robust cost function, and $\pi(\cdot,\cdot)$ is the projection function that projects $^{\mathcal{L}}p_{f_{i}}$ onto the image under pose $^{\mathcal{L}}x_{c_{i}}$, $u_{i,j}$ denotes the corresponding image feature point. $\Omega_{i,j}$ is the information matrix of the reprojection error. {\color{blue}{$E_{pl}(^{\mathcal{L}}p_{c_{j}})$ stands for the point-to-plane error term for $j$-th feature point and $v$-th laser point.}}
\begin{equation}\label{eq.sub7}
  E_{pl}(^{\mathcal{L}}p_{c_{j}})=\rho((n_{r_{v}}^{T}({^{\mathcal{L}}p_{c_{j}}}-p_{r_{v}}))^{T}\Omega_{j}(n_{r_{v}}^{T}({^{\mathcal{L}}p_{c_{j}}}-p_{r_{v}})))
\end{equation}
where $\Omega_{j}$ is the information matrix of the point-to-plane error.

We solve this optimization problem with the Gauss–Newton algorithm implemented in g2o \cite{kummerle2011g}. After the optimization we get the resulting calibrated extrinsic parameters.

\section{Observability Analysis}
In this work, we propose a LiDAR–camera calibration method based on 3D SLAM and discuss the observability of the system. Unlike the general SLAM system, our system adds new observations that will change the system's observability. In contrast to the observability analysis of LiDAR-Ladybug calibration \cite{mirzaei20123d}, our method uses a camera to obtain continuous images to reconstruct the 3D point cloud and analyzes the observability from the perspective of the dynamic system. {\color{blue}{The idea of observability analysis of our calibration system is as follows: comparing with the observation in general SLAM system, our system implement a new point-to-plane error measurement (i.e. $h_{pl}$). So we begin from the observability of the general SLAM system, and extend the analysis to our system. The observability is analyzed by determining the rank de-efficient of the observability matrix, i.e., the unobservable directions. Therefore, we focus on the nullspace of the observability matrix which describes the unobservable directions of the state space for which no information is provided by the measurement. Then we substitute the nullspace of the general SLAM system into our calibration system which has one more type of measurement (i.e. $h_{pl}$), and examine whether the dimensions remain unobservable.}}

\subsection{Observability Of Standard SLAM}
{\color{blue}{We follow the observability analysis of \cite{li2013high} about 3D Extended Kalman Filter (EKF)-based Visual-Inertial Odometry (VIO) and their derivation about rotation-error propagation equation.}} However, our position-error propagation equation is different from \cite{li2013high}, and there is no inertial measurement unit (IMU) in our system. {\color{blue}{In our analysis, the laser frame $\{\mathcal{L}\}$ is set as global frame of SLAM system,}} and the pose of camera $^{\mathcal{L}}x_{c}$ is expressed as the translation and the rotation in quaternion. Thus, the state vector at time $t_{k}$ is given by\footnote{Throughout this paper, $\hat{x} $ is used to denote the estimate of random variable $x$, while $\tilde{x}=x-\hat{x}$ is the error estimate. $\lfloor c\times \rfloor$ denotes the skew symmetric matrix corresponding to vector $c$. {\color{blue}{And $\breve{\cdot}$ is used to denote the matrix of general SLAM system.}}}:
\begin{equation}\label{eq.sub8}
  x_{k}=[\begin{matrix}^{\mathcal{L}}x_{c_{k}}^{T} & ^{\mathcal{L}}p_{f_{k}}^{T}\end{matrix}]^{T}=[\begin{matrix}_{\mathcal{L}}^{c_{k}}\bar{q}^{T} & ^{\mathcal{L}}p_{k}^{T} & ^{\mathcal{L}}p_{f_{k}}^{T}\end{matrix}]^{T}
\end{equation}
where $_{\mathcal{L}}^{c_{k}}\bar{q}$ is the unit quaternion representing the rotation from the laser frame to the camera frame at time $t_{k}$, $^{\mathcal{L}}p_{k}$ and ${^{\mathcal{L}}p_{f_{k}}}$ are the camera position and landmark position in the laser frame. As for the error state of position, we exploit the standard additive error definition (e.g.\, $^{\mathcal{L}}\tilde{p_{k}}={^{\mathcal{L}}p_{k}}-{^{\mathcal{L}}\hat{p_{k}}}$). And we define the rotation error based on the quaternion $\delta \bar{q}$:
\begin{equation}\label{eq.sub9}
  _{\mathcal{L}}^{c_{k}}\bar{q}={_{\mathcal{L}}^{c_{k}}\hat{\bar{q}}} \otimes \delta \bar{q}\Rightarrow\delta \bar{q}={_{\mathcal{L}}^{c_{k}}\hat{\bar{q}}}^{-1} \otimes {_{\mathcal{L}}^{c_{k}}\bar{q}}
\end{equation}
where $\otimes$ denotes quaternion multiplication, $\delta \bar{q}$ is a small rotation used to transform the estimated laser frame to match with the true one. As proposed in \cite{li2013high}, we rewrite $\delta \bar{q}$ to obtain a minimal 3-dimensional representation for this rotation:
\begin{equation}\label{eq.sub10}
  \delta \bar{q}=\begin{bmatrix} \frac{1}{2}^{\mathcal{L}}\tilde{\theta} \\ \sqrt{1-\frac{1}{4}{^{\mathcal{L}}\tilde{\theta}^{T}}\cdot{^{\mathcal{L}}\tilde{\theta}}}\end{bmatrix}\simeq\begin{bmatrix} \frac{1}{2}^{\mathcal{L}}\tilde{\theta} \\ 0 \end{bmatrix}
\quad
\end{equation}
{\color{blue}{where $^{\mathcal{L}}\tilde{\theta}$ is a $3\times1$ vector describing the rotation errors about the three axes. With the above error definition, the error-state is defined as:}}
\begin{equation}\label{eq.sub11}
  \tilde{x}=[\begin{matrix}^{\mathcal{L}}\tilde{\theta}^{T} & ^{\mathcal{L}}\tilde{p}^{T} & ^{\mathcal{L}}\tilde{p}_{f}^{T}\end{matrix}]^{T}
\end{equation}

We now turn attention to computing the error-state transition matrix. We note that the rotation-error definition satisfies:
\begin{equation}\label{eq.sub12}
  {_{\mathcal{L}}^{c_{k}}R}={_{\mathcal{L}}^{c_{k}}\hat{R}}(I_{3}-\lfloor ^{\mathcal{L}}\tilde{\theta}_{k}\times \rfloor)
\end{equation}
where ${_{\mathcal{L}}^{c_{k}}R}$ is the rotation between camera and laser frame, {\color{blue}{$^{\mathcal{L}}\tilde{\theta}_{k}$ is the $3\times1$ rotation vector at time $t_{k}$}}. And in the time interval $\begin{bmatrix} t_{k},t_{k+1} \end{bmatrix}$, rotation-error satisfies:
\begin{equation}\label{eq.sub13}
  {_{c_{k}}^{c_{k+1}}\bar{q}}={_{c_{k}}^{c_{k+1}}\hat{\bar{q}}}\otimes\delta \bar{q}_{\Delta t}
\end{equation}

From (\ref{eq.sub12}), we can obtain:
\begin{equation}\label{eq.sub14}
  {_{c_{k}}^{c_{k+1}}R}={_{c_{k}}^{c_{k+1}}\hat{R}}\cdot(I_{3}-\lfloor ^{\mathcal{L}}\tilde{\theta}_{\Delta t}\times \rfloor)
\end{equation}
{\color{blue}{where $^{\mathcal{L}}\tilde{\theta}_{\Delta t}$ is a $3\times1$ error vector.}} Therefore, we can obtain the linearized expression for the rotation-error propagation:
\begin{equation}\label{eq.sub15}
  ^{\mathcal{L}}\tilde{\theta}_{k+1}\simeq {^{\mathcal{L}}\tilde{\theta}_{k}}+\hat{R}_{k}^{T}\tilde{\theta}_{\Delta t}
\end{equation}
where $\hat{R}_{k}$ is the shorthand notation of ${_{\mathcal{L}}^{c_{k+1}}\hat{R}}$. Additionally, the position propagation equation is:
\begin{equation}\label{eq.sub16}
  {^{\mathcal{L}}\hat{p}_{k+1}}\simeq \hat{R}_{k}^{T}\cdot{^{k}\hat{p}_{k+1}}+{^{\mathcal{L}}\hat{p}_{k}}
\end{equation}

In order to get the linearized position-error propagation, we use (\ref{eq.sub12}) to linearize (\ref{eq.sub16}):
\begin{equation}\label{eq.sub17}
  {^{\mathcal{L}}\tilde{p}_{k+1}}\simeq -\lfloor (\hat{R}_{k}^{T}\cdot{^{k}\hat{p}_{k+1}})\times\rfloor{^{\mathcal{L}}\tilde{\theta}_{k}}+\hat{R}_{k}^{T}\cdot{^{k}\tilde{p}_{k+1}}+{^{\mathcal{L}}\tilde{p}_{k}}
\end{equation}
where ${^{k}\tilde{p}_{k+1}}={^{k}p_{k+1}}-{^{k}\hat{p}_{k+1}}$ denotes the error in ${^{k}\hat{p}_{k+1}}$. Since the landmark is static, its state estimate is invariant with time. Similarly, the landmark-error propagation is derived:
\begin{equation}\label{eq.sub18}
  {^{\mathcal{L}}\tilde{p}_{f_{k+1}}}={^{\mathcal{L}}\tilde{p}_{f_{k}}}
\end{equation}

{\color{blue}{According to (\ref{eq.sub15}), (\ref{eq.sub17}) and (\ref{eq.sub18}), we can obtain the error-state propagation equation:}}
\begin{equation}\label{eq.sub19}
\begin{split}
   \underbrace{{\begin{bmatrix} ^{\mathcal{L}}\tilde{\theta}_{k+1} \\ {^{\mathcal{L}}\tilde{p}_{k+1}} \\ {^{\mathcal{L}}\tilde{p}_{f_{k+1}}} \end{bmatrix}}}_{\tilde{x}_{k+1}} = \underbrace{{\begin{bmatrix} I_{3} & 0_{3} & 0_{3\times3m} \\ -\lfloor (\hat{R}_{k}^{T}\cdot{^{k}\hat{p}_{k+1}})\times\rfloor & I_{3} & 0_{3\times3m} \\ 0_{3m\times3} & 0_{3m\times3} & I_{3m\times3m}\end{bmatrix}}}_{\breve{\phi}_{k}} \\
  \cdot \underbrace{{\begin{bmatrix} {^{\mathcal{L}}\tilde{\theta}_{k}} \\ {^{\mathcal{L}}\tilde{p}_{k}} \\ {^{\mathcal{L}}\tilde{p}_{f_{k}}} \end{bmatrix}}}_{\tilde{x}_{k}}
    +\underbrace{{\begin{bmatrix} \hat{R}_{k}^{T}\tilde{\theta}_{\Delta t} \\ \hat{R}_{k}^{T}\cdot{^{k}\tilde{p}_{k+1}} \\ 0_{3m\times3m} \end{bmatrix}}}_{w_{d_{k}}}
\end{split}
\end{equation}
where $m$ is the number of landmarks, $-\lfloor (\hat{R}_{k}^{T}\cdot{^{k}\hat{p}_{k+1}})\times\rfloor$ can be represented as $-\lfloor ({^{\mathcal{L}}\hat{p}_{k+1}}-{^{\mathcal{L}}\hat{p}_{k}})\times\rfloor$, $\breve{\phi}_{k}$ is the error-state transition matrix, $w_{d_{k}}$ is the noise process. Different from \cite{li2013high}, our error-state transition matrix has only three variables. {\color{blue}{The measurement error in general SLAM system at time $t_{k}$ for landmark feature $i$ is the feature reprojection measurement error $h_{proj}$ according to (\ref{eq.sub6}):}}
\begin{equation}\label{eq.sub2288}
  h_{proj}=\pi(^{\mathcal{L}}p_{f_{i}},^{\mathcal{L}}x_{c_{k}})-u_{i,k}
\end{equation}

Therefore the measurement Jacobian matrix $\breve{H}_{ik}$ is given by:
\begin{equation}\label{eq.sub20}
  \breve{H}_{ik}{\color{blue}{\triangleq \frac{\partial h_{proj}}{\partial x_{k}} }} =\begin{bmatrix} \breve{H}_{I_{ik}} & 0 & \ldots & \breve{H}_{f_{ik}} & \ldots & 0 \end{bmatrix}
\end{equation}
where the Jacobian of reprojection measurement with respect to the landmark feature position and camera pose are given by, separately:
\begin{equation}\label{eq.sub21}
  \breve{H}_{f_{ik}}{\color{blue}{\triangleq \frac{\partial h_{proj}}{\partial ^{\mathcal{L}}p_{f_{i}}  } }}=\breve{J}_{ik}\cdot\hat{R}_{k}
\end{equation}
\begin{equation}\label{eq.sub22}
  \breve{H}_{I_{ik}}{\color{blue}{\triangleq \frac{\partial h_{proj}}{\partial ^{\mathcal{L}}x_{c_{k}}} }}= \breve{H}_{f_{ik}}\begin{bmatrix} -\lfloor ({^{\mathcal{L}}\hat{p}_{f_{i}}}-{^{\mathcal{L}}\hat{p}_{k}})\times\rfloor & -I_{3} \end{bmatrix}
\end{equation}
{\color{blue}{where $\breve{J}_{ik}\triangleq \partial {h_{proj}}/{\partial ^{c_{k}}p_{f_{i}}}$ is the Jacobian of reprojection measurement with respect to the landmark feature position in camera frame at time $t_{k}$.}}

{\color{blue}{Then we analyze the observability matrix, since the nullspace of the observability matrix describes the directions of the state space for which no information is provided by the measurement, i.e., the unobservable directions. The observability matrix for the time interval between time $t_{s}$ and $t_{s+w}$ is defined following \cite{chen1990local} as:}}
\begin{equation}\label{eq.sub11123}
  \breve{M}\triangleq {\begin{bmatrix} \breve{M}_{s} & \ldots & \breve{M}_{k} & \ldots & \breve{M}_{s+w} \end{bmatrix}}^{T}
\end{equation}
{\color{blue}{where time $t_{k}$ is between time $t_{s}$ and $t_{s+w}$. And $\breve{M}_{k}$ is defined as:}}
\begin{equation}\label{eq.sub111111123}
  \breve{M}_{k}\triangleq {\begin{bmatrix} \breve{M}_{1k} & \ldots & \breve{M}_{ik} & \ldots \end{bmatrix}}^{T}
\end{equation}
{\color{blue}{where $\breve{M}_{ik}$ is the block row of the observability matrix corresponding to the observation of landmark feature $i$ at time $t_{k}$ between time $t_{s}$ and $t_{s+w}$, which is defined as:}}
\begin{multline}\label{eq.sub24} \check{M}_{ik}\triangleq
  {\color{blue}{\breve{H}_{ik}\breve{\phi}_{k-1}\cdots\breve{\phi}_{s}}}
  =\breve{J}_{ik}\hat{R}_{k}\left[\begin{array}{ccc}
  \breve{\Gamma}_{ik} & -I_{3} & 0_{3}
  \end{array}\right.\\
\left.\begin{array}{cccc}
  \ldots & I_{3} & \ldots & 0_{3} \\
  \end{array}\right]
\end{multline}
\begin{equation}\label{eq.sub25}
  \breve{\Gamma}_{ik}\triangleq\lfloor ({^{\mathcal{L}}{p}_{f_{i}}}-{^{\mathcal{L}}{p}_{s}})\times\rfloor
\end{equation}
where $\phi_{s}$ is the error-state transition matrix in time $t_{s}$.

At this point, we define the nullspace as follows:
\begin{equation}\label{eq.sub26}
  \breve{N}\triangleq\begin{bmatrix} 0_{3} & I_{3} \\ I_{3} & -\lfloor {^{\mathcal{L}}{p}_{s}}\times\rfloor \\ I_{3} & -\lfloor {^{\mathcal{L}}{p}_{f_{1}}}\times\rfloor \\ \vdots & \vdots \\ I_{3} & -\lfloor {^{\mathcal{L}}{p}_{f_{m}}}\times\rfloor \end{bmatrix}
\end{equation}

It is easy to verify that $\breve{M}_{ik}\cdot\breve{N}=0_{2\times6}$. Since this holds for any $i$ and any $k$ (i.e.\, for all block rows of the observability matrix), we conclude that $\breve{M}\cdot\breve{N}=0$. {\color{blue}{That is to say, the general 3D SLAM system is unobservable. It means that only the 6DoF pose of the current time relative to the initial time can be determined, which is explained as the global pose at the initial time of the general SLAM system (i.e. ${^{\mathcal{L}}x_{c_{0}}}$, which is in $x_{0}$ defined in (\ref{eq.sub8}), is the $_{\mathcal{C}}^{\mathcal{L}}x$ in our calibration system) is unobservable. That is to say, it is impossible to calibrate the extrinsic parameters by the general SLAM system. So in Section III-B, our focus is to analyze whether the observability of our calibration system will change when new point-to-plane error measurement (i.e. $h_{pl}$) is added.}}

\subsection{Observability Of Our Calibration System}
{\color{blue}{Comparing with the observation in general SLAM system, we find that in our calibration system the one more type of measurement changes the measurement Jacobian matrix and the observability matrix. The error-state transition matrix is the same as the standard one. Turning to the measurement Jacobian matrix, besides the feature reprojection measurement error $h_{proj}$, there is point-to-plane measurement error $h_{pl}$ in our system according to (\ref{eq.sub7}):}}
\begin{equation}\label{eq.sub228}
  h_{pl}=n_{r}^{T}({^{\mathcal{L}}p_{c}}-p_{r})
\end{equation}

{\color{blue}{Thus, the measurement Jacobian matrix $H_{ik}$ at time $t_{k}$ for landmark feature $i$ is given by:}}
\begin{equation}\label{eq.sub28}
  {H}_{ik}\triangleq\begin{bmatrix} \breve{H}_{ik} \\ H_{pl} \end{bmatrix}=\begin{bmatrix} \breve{H}_{I_{ik}} & 0 & \ldots & \breve{H}_{f_{ik}} & \ldots & 0 \\ 0 & 0 & \ldots & n_{r}^{T} & \ldots & 0\end{bmatrix}
\end{equation}
\begin{equation}\label{eq.sub22222228}
  H_{pl}{\color{blue}{\triangleq \frac{\partial h_{pl}}{\partial x_{k}} }}=\begin{bmatrix} 0 & 0 & \ldots & n_{r}^{T} & \ldots & 0\end{bmatrix}
\end{equation}
{\color{blue}{where $H_{pl}$ refers to the Jacobian matrix of $h_{pl}$ with respect to $^{\mathcal{L}}x_{c}$ and ${^{\mathcal{L}}p_{f}}$ (the point belonging to the chessboard is represented as $^{\mathcal{L}}p_{c}$). Then according to (\ref{eq.sub24}), the observability matrix of feature $i$ at time $t_{k}$ becomes:}}
\begin{equation}\label{eq.sub29}
  {M}_{ik}\triangleq
  \begin{bmatrix}
  \breve{J}_{ik}\hat{R}_{k}\begin{bmatrix} \breve{\Gamma}_{ik} & -I_{3} & 0_{3} & \ldots & I_{3} & \ldots & 0_{3} \end{bmatrix} \\
  \begin{bmatrix} 0 & 0 & 0 & \ldots & n_{r}^{T} & \ldots & 0   \end{bmatrix}
  \end{bmatrix}
\end{equation}
{\color{blue}{where $I_{3}$ and $n_{r}^{T}$ are both in the $i-th$ column of the matrix.}}

Now we analyze the minimal necessary conditions of the number of calibration targets to solve the accurate 6DoF extrinsic calibration problem.

\textbf{Observation of one plane:} Suppose there are enough points on each chessboard plane, the observability matrix of chessboard's feature $i$ and feature $j$ at time $t_{k}$ becomes:
\begin{equation}\label{eq.sub300}
  {M}_{(i,j)k}\triangleq
  \begin{bmatrix}
  \breve{J}_{ik}\hat{R}_{k}\begin{bmatrix} \breve{\Gamma}_{ik} & -I_{3} & 0_{3} & \ldots & I_{3} & \ldots & 0_{3} \end{bmatrix} \\
  \breve{J}_{jk}\hat{R}_{k}\begin{bmatrix} \breve{\Gamma}_{jk} & -I_{3} & 0_{3} & \ldots & \ldots & I_{3} & 0_{3} \end{bmatrix} \\
  \begin{bmatrix} 0 & 0 & 0 & \ldots & n_{r}^{T} & \ldots & 0   \end{bmatrix}\\
  \begin{bmatrix} 0 & 0 & 0 & \ldots & \ldots & n_{r}^{T} & 0   \end{bmatrix}
  \end{bmatrix}
\end{equation}

{\color{blue}{The derivation of the nullspace of the observability matrix is given in Appendix-A. We follow the stacking in (\ref{eq.sub11123}) to arrange $M_{(i,j)k}$, thus forming $M$. Note that there exists:
\begin{equation}\label{eq.sub22222222222229}
  M_{(i,j)k}N_{1}=0_{6\times3}
\end{equation}
where $N_{1}$ is described in Appendix-A.

Since this holds for any $i$, $j$ and any $k$, we conclude that $MN_{1}=0$. Therefore, when observing only one plane, any translation parallel to the plane's normal and any rotation around the plane's normal vector is unobservable.}}

\textbf{Observation of two planes:} the observability matrix of feature $i$ and feature $j$ at time $t_{k}$ from two chessboards, described by $n_{r_{a}}$,and $n_{r_{b}}$:
\begin{equation}\label{eq.sub344}
  {M}_{(i,j)k}\triangleq
  \begin{bmatrix}
  \breve{J}_{ik}\hat{R}_{k}\begin{bmatrix} \breve{\Gamma}_{ik} & -I_{3} & 0_{3} & \ldots & I_{3} & \ldots & 0_{3} \end{bmatrix} \\
  \breve{J}_{jk}\hat{R}_{k}\begin{bmatrix} \breve{\Gamma}_{jk} & -I_{3} & 0_{3} & \ldots & \ldots & I_{3} & 0_{3} \end{bmatrix} \\
  \begin{bmatrix} 0 & 0 & 0 & \ldots & n_{r_{a}}^{T} & \ldots & 0   \end{bmatrix}\\
  \begin{bmatrix} 0 & 0 & 0 & \ldots & \ldots & n_{r_{b}}^{T} & 0   \end{bmatrix}
  \end{bmatrix}
\end{equation}

{\color{blue}{Note that the third and forth row in (\ref{eq.sub344}) are different from that in (\ref{eq.sub300}). For this block of observability matrix, we have
\begin{equation}\label{eq.sub222222222222229}
  M_{(i,j)k}N_{2}=0_{6\times1}
\end{equation}
where $N_{2}$ is described in Appendix-A. Since this holds for any $i$, $j$ and any $k$, we conclude that $MN_{2}=0$. Therefore, when observing two planes, one degree of freedom of the translation is unobservable.}}

\textbf{Observation of three planes:} {\color{blue}{Similar to the previous derivation process, we conclude that when three planes with non-collinear normal vectors are observed, we can determine all the unknowns. That is to say, our calibration system is observable. The above observability analysis proves that at least three chessboards are needed to calibrate the 6DoF extrinsic parameters. In other words, a larger common field of view between the two sensors is needed to guarantee the observability and reliable detection of the calibration targets, which is difficult for the appearance-based calibration method.}}

{\color{blue}{In order to enrich our theory and make it suitable for different calibration targets, we have added Appendix-B: ``observability of our calibration system with point-to-point error measurement''. Unlike the chessboards, which represent as the normal vector to construct the point-to-plane error (i.e. $h_{pl}$), some calibration targets, i.e. polygonal planar boards \cite{park2014calibration} and boxes \cite{pusztai2017accurate}, are detected with corner points, which leads to point-to-point error measurement. The result of the observability analysis to this class of calibration targets is that when observing only one point, any rotation is unobservable, while observing two points, one degree of freedom of the rotation is unobservable. Also when three non-collinear points are observed, we can determine all the unknowns.}}

\section{Placement Of Calibration Targets}
The visual information in this method is represented by a point cloud in space of a set of sparse visual 3D chessboards. The previous analysis concluded that in order to calibrate the 6DoF extrinsic parameters, 3 chessboards were needed at least. How to place these 3 chessboards in space to get the most comprehensive visual information and better calibration accuracy is what we will discuss next. In the appearance-based method, the increase in the number of chessboards has little effect on the calibration accuracy, so we investigate the effect of the placement of the chessboards on the calibration accuracy. The following theory can provide a guideline for designing high-accuracy calibration procedures.

To simplify the problem we are analyzing, we consider that the visual 3D points reconstructed by our method are one frame of visual data. We derive the problem in 2D and consider each two chessboards which can still provide insights into real-world applications. Analyzing the calibration accuracy refers to analyzing the uncertainty of the point-to-plane error function, which can be represented as a determination of the corresponding Hessian matrix. The larger the determinant of the Hessian matrix, the smaller the uncertainty. Specifically, we are going to explore the influence of the angle and distance between two chessboards respectively.

\subsection{Angle Between Calibration Targets}
\begin{figure}[tp]
     \centering
      \includegraphics[scale=0.26]{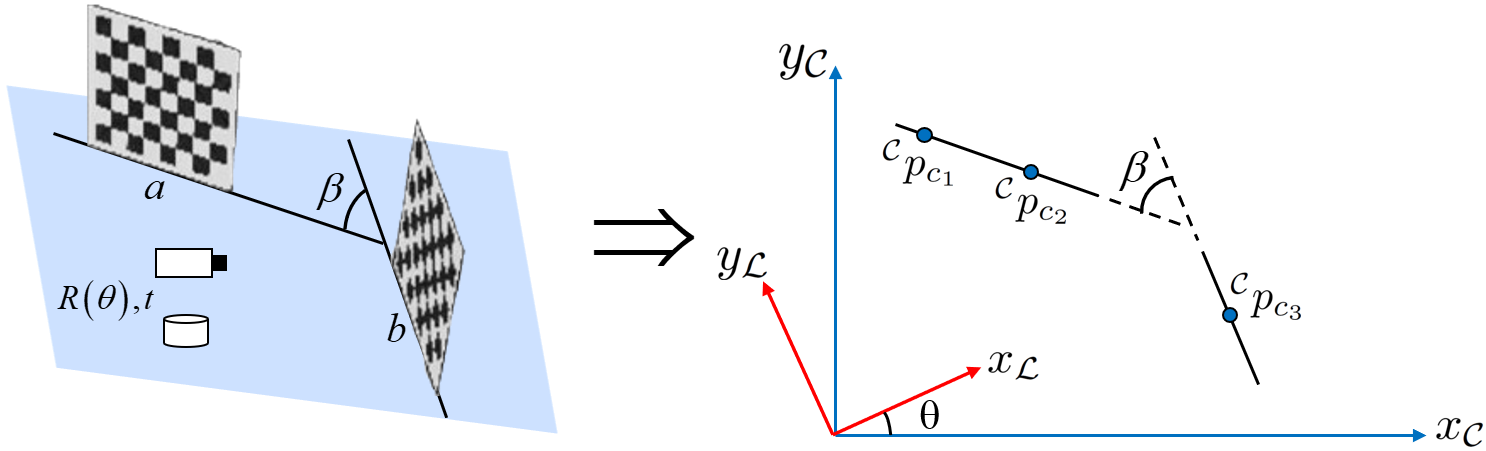}
      \caption{Specific explanation of the parameters and representation of 2D.}
      \label{explanation}
\end{figure}

As shown in Fig. \ref{explanation}, the angle of chessboard $b$ relative to chessboard $a$ is $\beta$, the angle of $\{\mathcal{C}\}$ relative to $\{\mathcal{L}\}$ is $\theta$, and $_{\mathcal{C}}^{\mathcal{L}}x$ is reduced to 3DoF represented by
\begin{equation}\label{eq.sub39}
  R(\theta)=\begin{bmatrix} \cos\theta & -\sin\theta \\ \sin\theta & \cos\theta \end{bmatrix} , t=\begin{bmatrix} t_{x} \\ t_{y} \end{bmatrix}
\end{equation}

Under the simplified condition, the point-to-plane error item can be rewritten as:
\begin{equation}\label{eq.sub40}
  \varepsilon_{pl}=(n_{r}^{T}\cdot(R(\theta)\cdot {^{\mathcal{C}}p_{c}}+t-p_{r}))
\end{equation}
where $^{\mathcal{C}}p_{c}$ is the visual 2D point in the camera coordinate system, $R(\theta)$ and $t$ are the 2D extrinsic parameters and $p_{r}$ is the laser 2D point in the laser coordinate system, $n_{r}$ is the 2D normal vector of $p_{r}$.
Since the Gauss–Newton method \cite{bjorck1996numerical} is one of the simplest and most versatile methods for optimization algorithms, our proof uses this method to solve the gradient descent direction. The Gauss–Newton method uses $J^{T}J$ as an approximation of the second-order Hessian matrix. For our system, the Jacobian matrix of the error function relative to the extrinsic parameters $R(\theta)$, $t$ is derived as follows:
\begin{equation}\label{eq.sub41}
  J\triangleq\begin{bmatrix} n_{r}^{T}\cdot({R(\theta)'}\cdot {^{\mathcal{C}}p_{c}}) & n_{r}^{T} \end{bmatrix}
\end{equation}
where ${R(\theta)'}$ denotes the derivatives of ${R(\theta)}$ with respect to $\theta$. Further, the Hessian matrix is as follows:
\begin{equation}\label{eq.sub42}
  H\triangleq J^{T}\cdot J
\end{equation}

In order to explore the placement of the calibration plate, we have further simplified the situation. Assuming that the normal vector of the calibration plate $a$ is
\begin{equation}\label{eq.sub43}
  n_{1}\triangleq\begin{bmatrix} 1 & 0 \end{bmatrix}^{T}
\end{equation}

Then the normal vector of the calibration plate $b$ is
\begin{equation}\label{eq.sub44}
  n_{2}\triangleq\begin{bmatrix} \cos\beta & \sin\beta \end{bmatrix}^{T}
\end{equation}

Thereafter, we take two points ${^{\mathcal{C}}p_{c_{1}}}$ and ${^{\mathcal{C}}p_{c_{2}}}$ from the calibration plate $a$, and one point ${^{\mathcal{C}}p_{c_{3}}}$ from the calibration plate $b$. Then, for these three pairs of points, the Hessian matrix of the error function that needs to be jointly optimized is:
\begin{equation}\label{eq.sub45}
\begin{split}
   H\triangleq J_{1}^{T}J_{1}+J_{2}^{T}J_{2}+J_{3}^{T}J_{3} \\
   \quad
   J_{1}\triangleq\begin{bmatrix} n_{1}^{T}\cdot({R(\theta)'}\cdot {^{\mathcal{C}}p_{c_{1}}}) & n_{1}^{T} \end{bmatrix} \\
   \quad
   J_{2}\triangleq\begin{bmatrix} n_{1}^{T}\cdot({R(\theta)'}\cdot {^{\mathcal{C}}p_{c_{2}}}) & n_{1}^{T} \end{bmatrix} \\
   \quad
   J_{3}\triangleq\begin{bmatrix} n_{2}^{T}\cdot({R(\theta)'}\cdot {^{\mathcal{C}}p_{c_{3}}}) & n_{2}^{T} \end{bmatrix}
\end{split}
\end{equation}

Finding the determinant of the Hessian matrix and simplifying it we can obtain:
\begin{equation}\label{eq.sub46}
  |H|=(Q_{c_{1y}}-Q_{c_{2y}})^{2}\cdot\sin^{2}(\beta)
\end{equation}
{\color{blue}{where ${^{\mathcal{C}}p_{c_{1}}}\triangleq\begin{bmatrix} p_{c_{1_{x}}} & p_{c_{1_{y}}} \end{bmatrix}^{T}$,  ${^{\mathcal{C}}p_{c_{2}}}\triangleq\begin{bmatrix} p_{c_{2_{x}}} & p_{c_{2_{y}}} \end{bmatrix}^{T}$, and $Q_{c_{1y}}\triangleq\cos\theta\cdot{p_{c_{1_{x}}}}-\sin\theta\cdot{p_{c_{1_{y}}}}$, $Q_{c_{2y}}\triangleq\cos\theta\cdot{p_{c_{2_{x}}}}-\sin\theta\cdot{p_{c_{2_{y}}}}$.}}

It can be seen that when $\beta=\frac{\pi}{2}$, $|H|$ takes the maximum value, that is, when the two plates are at $90^{\circ}$, the uncertainty is the smallest. Obviously, it is difficult to observe the obverse sides of the above two chessboards at the same time for the common field of view provided by the appearance-based method.
\begin{figure}[tp]
     \centering
      \includegraphics[scale=0.3]{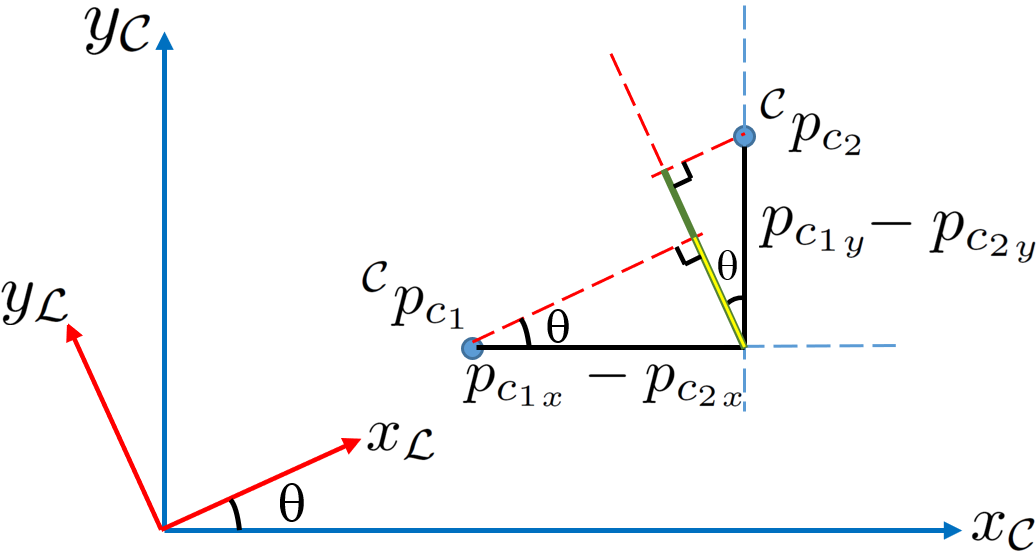}
      \caption{Schematic diagram of (\ref{eq.sub51}), the green line is the result of projection.}
      \label{schematic}
\end{figure}
\subsection{Distance Between Calibration Targets}
The conclusion of Section IV-A is that the calibration error uncertainty is the smallest when the two chessboards are placed orthogonal to each other. In this section, when the angles of the two chessboards are fixed, we will discuss the effect of the distance between two chessboards.

However, the distance between the two chessboards is not directly involved in the optimization; in fact it is the points selected on the chessboard that are pertinent to the optimization. Therefore, the problem is equal to how the distance between the points influences the calibration accuracy, based on the premise that the existence of measurement errors is not considered in our derivation.

Again, we simplify the problem to 2D situation. One should notice that the reference frame of the normal vectors is the laser frame. Assume that the angle of calibration plate $a$ relative to the $x$-axis of $\{\mathcal{L}\}$ is $\alpha$, and the angle of calibration plate $b$ with respect to the $x$-axis of $\{\mathcal{L}\}$ is $\alpha+\beta$. Combining the conclusions of Section IV-A, and to simplify the derivation, we set $\alpha=0$ and $\beta=\frac{\pi}{2}$. Therefore, the normal vector of calibration plate $a$ is:
\begin{equation}\label{eq.sub47}
  n_{1}\triangleq\begin{bmatrix} 1 & 0 \end{bmatrix}^{T}
\end{equation}

The normal vector of calibration plate $b$ is:
\begin{equation}\label{eq.sub48}
  n_{2}\triangleq\begin{bmatrix} 0 & 1 \end{bmatrix}^{T}
\end{equation}

As before, we take two points ${^{\mathcal{C}}p_{c_{1}}}$ and ${^{\mathcal{C}}p_{c_{2}}}$ from calibration plate $a$ and one point ${^{\mathcal{C}}p_{c_{3}}}$ from calibration plate $b$. The Jacobian matrix of the error function relative to the extrinsic parameters $R(\theta)$, $t$ is as follows:
\begin{equation}\label{eq.sub49}
  J\triangleq\begin{bmatrix} n_{r}^{T}\cdot({R(\theta)'}\cdot {^{\mathcal{C}}p_{c}}) & n_{r}^{T} \end{bmatrix}
\end{equation}

Then, the Hessian matrix is:
\begin{equation}\label{eq.sub50}
  \begin{split}
   H\triangleq J_{1}^{T}J_{1}+J_{2}^{T}J_{2}+J_{3}^{T}J_{3} \\
   \quad
   J_{1}\triangleq\begin{bmatrix} n_{1}^{T}\cdot({R(\theta)'}\cdot {^{\mathcal{C}}p_{c_{1}}}) & n_{1}^{T} \end{bmatrix} \\
   \quad
   J_{2}\triangleq\begin{bmatrix} n_{1}^{T}\cdot({R(\theta)'}\cdot {^{\mathcal{C}}p_{c_{2}}}) & n_{1}^{T} \end{bmatrix} \\
   \quad
   J_{3}\triangleq\begin{bmatrix} n_{2}^{T}\cdot({R(\theta)'}\cdot {^{\mathcal{C}}p_{c_{3}}}) & n_{2}^{T} \end{bmatrix}
\end{split}
\end{equation}

Solving the determinant of the Hessian matrix and simplifying it, we can get
\begin{equation}\label{eq.sub51}
   |H|=((p_{c_{1_{x}}}-p_{c_{2_{x}}})\cdot\sin\theta+(p_{c_{1_{y}}}-p_{c_{2_{y}}})\cdot\cos\theta)^{2}
\end{equation}
where ${^{\mathcal{C}}p_{c_{1}}}\triangleq\begin{bmatrix} p_{c_{1_{x}}} & p_{c_{1_{y}}} \end{bmatrix}^{T}$,  ${^{\mathcal{C}}p_{c_{2}}}\triangleq\begin{bmatrix} p_{c_{2_{x}}} & p_{c_{2_{y}}} \end{bmatrix}^{T}$, and $((p_{c_{1_{x}}}-p_{c_{2_{x}}})\cdot\sin\theta)$ is the value projected into the $y$-axis of the laser coordinate system. $((p_{c_{1_{y}}}-p_{c_{2_{y}}})\cdot\cos\theta)$ is also a value projected into the $y$-axis of the laser coordinate system. The illustration of (\ref{eq.sub51}) can be seen in Fig. \ref{schematic}.

\begin{figure}[tbp]
     \centering
            \subfigure{
                \includegraphics[scale=0.8]{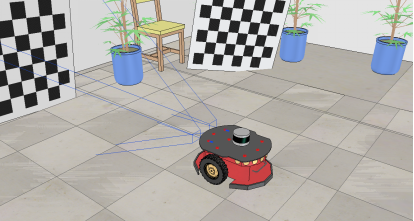}
            }\\
            \subfigure{
                \includegraphics[scale=0.35]{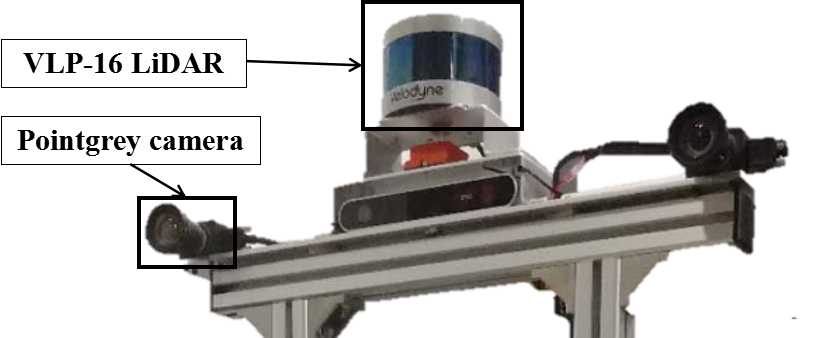}
            }
	\caption{Sensor configuration Top: VLP-16 LiDAR and vision sensor in the simulation environment.  Bottom: VLP-16 LiDAR  and Pointgrey camera in the real-world.}
	\label{sensor}	
\end{figure}

Therefore, we can find that the farther the projection distance on the $y$-axis is, the smaller the calibration error uncertainty is. Furthermore, since the size of the board cannot be infinitely large, multiple calibration plates are necessary to improve the calibration accuracy. When there are multiple calibration plates in the space, the angle between each two calibration plates should be as close as  possible to $90^{\circ}$ for better calibration accuracy in 2D. Therefore, the pose between each two calibration plates should be as different as possible to reduce the calibration error.

\textbf{Comments}

\begin{itemize}
  \item Combining the analysis of observability and the minimal necessary conditions for calibration we conclude that at least three chessboards are required and more chessboards can lead to better calibration accuracy.
  \item With the same number of calibration targets, a scattered placement is better than a centralized one, which is expected to be true in 3D.
  \item The extended camera field of view obtained by our method meets the requirement of observing multiple calibration targets, which is difficult in those methods that keep the sensors stationary.
  \item Observing multiple calibration targets arranged in various poses in the extended the field of view also gives our method an advantage compared to the appearance-based method.
\end{itemize}

\begin{figure}[tbp]
     \centering
      \includegraphics[scale=0.55]{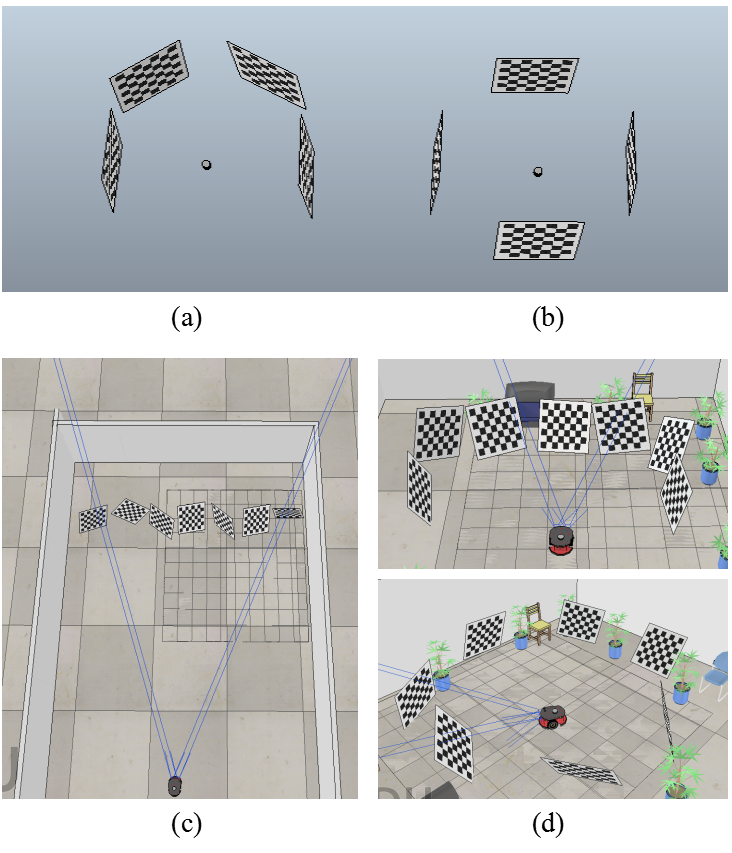}
      \caption{{\color{blue}{Scenes for obtaining data with chessboards}}. (a) and (b): {\color{blue}{Obtaining data in the scene in which four chessboards are placed nearly vertical to the ground.}} (c): The placement of the chessboards and the sensors to collect the data required by the KITTI single shot method. (d): Obtaining data in the scene in which seven chessboards are placed in various poses.}
      \label{Scenes}
\end{figure}
\section{Experimental Results}
In order to evaluate our method, we performed simulation verification and real-world experiments separately. In the simulation experiments, we showed that the placement theory derived in Section IV is reasonable by performing our calibration method with calibration targets placed in scattered and centralized arrangements, respectively. Then compared with the other methods, the results show the advantage of the proposed method on the calibration accuracy. {\color{blue}{We then used polygonal planar boards and boxes as calibration targets to examine the consistency of the proposed results when other types of measurements was applied.}} With regard to the evaluation criteria, the simulation environment provides the ground truth for the LiDAR and the camera's extrinsic parameters, so we could directly evaluate the calibration accuracy. We built a simulation environment in V-REP, using a stereo vision sensor and a Velodyne VLP-16 LiDAR to obtain data as shown in the top of Fig. \ref{sensor}.

In the real-world experiments, the comparison with other methods is also performed demonstrating the practicality of the proposed method. With regard to the evaluation criteria, the ground truth of the extrinsic parameters is not available. In order to evaluate the result of the estimated extrinsic parameters, we divided the collected data into two groups, one for training the point-to-plane error with different extrinsic results obtained by different methods, and the other for testing the accuracy. As shown in the bottom of Fig. \ref{sensor}, we fixed two Pointgrey cameras with a Velodyne VLP-16 LiDAR on the robot to perform the real-world experiments and the cameras were calibrated by default.

\begin{figure}[tp]
     \centering
      \includegraphics[scale=0.31]{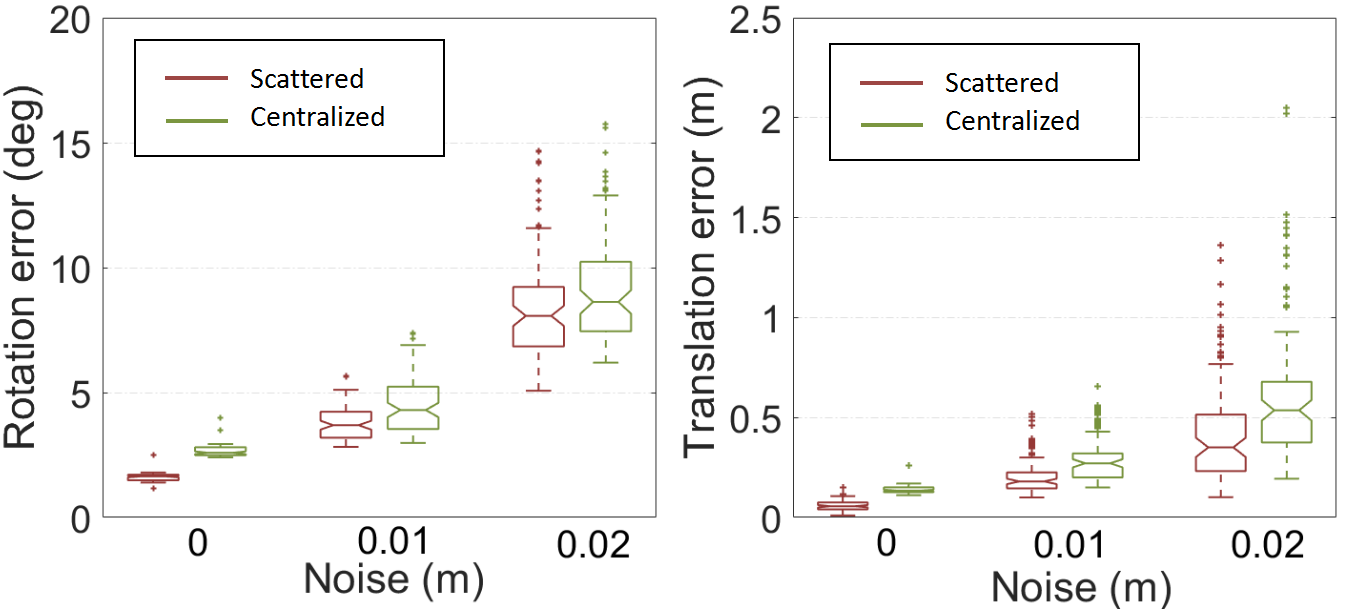}
      \caption{Theoretical verification results: errors from the ground truth of the calibration result by our method with chessboards placed centralized and scattered when adding Gaussian noise to the laser data.}
      \label{2D}
\end{figure}
\subsection{V-REP Simulation}
\subsubsection{Theoretical Verification}
In the simulation environment, we first verify the theoretically derived conclusions. Due to the existence of observation errors, the angle between the calibration targets and the distance between the calibration targets are highly coupled, and it is impossible to perform strict control variables to verify the influence of angle and distance separately. We only verify the final conclusions derived from the theory. As shown in Fig. \ref{Scenes} (a) and (b) we placed four chessboards around the sensor, and used our method to obtain data for calibration. {\color{blue}{In order to simulate the situation in 2D four chessboards are placed nearly vertical to the ground, and one set is centralized in front of the field of view, the other is scattered around the sensor.}} For providing sufficient constraints for calibration, we made the experiment in which the chessboards are placed at a 5 degree angle to the direction of gravity.

The final calibration $_{\mathcal{C}}^{\mathcal{L}}x$ is expressed as rotation $R$, and translation $t$. $R$, $t$ were compared against ground truth $R_{g}$, $t_{g}$, which were obtained from V-REP. Following \cite{geiger2012automatic}, for translation error, we computed $\|t-t_{g}\|$ in meters. For rotation error, we first computed the relative rotation $\delta R=R^{-1}R_{g}$ and represented it in degrees. The results, depicted in Fig. \ref{2D}, indicate that the calibration accuracy is better when the angle between the two chessboards is $90^{\circ}$ as shown in Fig. \ref{Scenes} (b), which is consistent with the conclusion of Section IV. As shown in Fig. \ref{Scenes} (a) and (b), in order to obtain a larger common field of view between the two sensors to observe four chessboards, the camera's extended field of view obtained by our method is needed.
\begin{figure}[tp]
     \centering
      \includegraphics[scale=0.45]{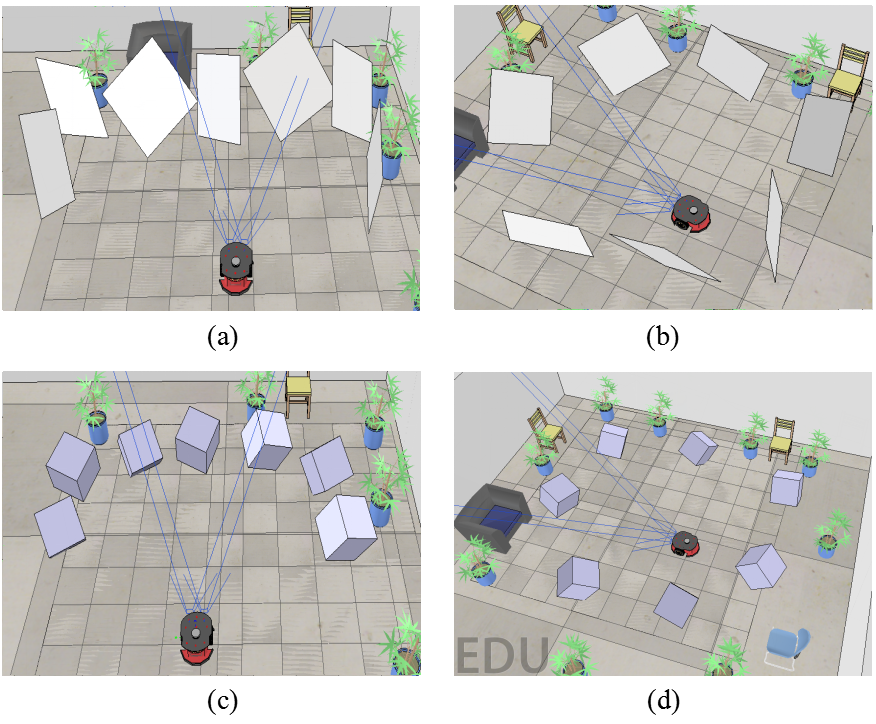}
      \caption{{\color{blue}{Scenes for obtaining data with polygonal planar boards and boxes.}}}
      \label{box}
\end{figure}
\subsubsection{Accuracy Comparison}
Next, we compared our method with the \textbf{KITTI single shot method} \cite{geiger2012automatic} on calibration accuracy. We placed seven chessboards in various poses as shown in Fig. \ref{Scenes} (c), and obtained data for the KITTI single shot calibration method to calibrate. The KITTI single shot calibration method can automatically give the extrinsic calibration results in one acquisition, which is convenient to use. The method requires placing multiple chessboards in front of the field of view and obtaining the LiDAR and camera data in one shot, respectively. However, if one wants to obtain as much data as possible from the chessboards, the sensor should be placed much farther away from the chessboards which can be easily seen from Fig. \ref{Scenes} (c). Once the sensor is too far from the chessboards, it is often difficult to extract the corner points from the obtained camera image, and the laser lines hitting the chessboards are also reduced. Another main limiting assumption of the KITTI approach is the common field of view between the camera and the LiDAR.

As shown in Fig. \ref{Scenes} (d) top and bottom, we then obtained two sets of data for seven chessboards centralized and scattered in a certain field of view and evaluated \textbf{proposed method} (using chessboards as an example). Unlike the KITTI single shot method, when we captured the camera images, we moved the robot around nearly to obtain images of each chessboard, and then reconstructed the global visual 3D points in the space. {\color{blue}{Our method does not limit the calibration target, so we used polygonal planar boards \cite{park2014calibration} (labeled as \textbf{polygonal method} as shown in Fig. \ref{box} (a) and (b)) and boxes \cite{pusztai2017accurate} (labeled as \textbf{box method} as shown in Fig. \ref{box} (c) and (d)) as calibration targets for a complement to our work. As the detection of the corner is not the point in this manuscript, for the LiDAR point clouds, we obtained the corner points of the polygonal planar boards or the calibration boxes from V-REP and manually add noise; and for the image data, we extracted points by Harris \cite{harris1988combined} corner detector and reconstructed them to the visual 3D points. After data association between these point clouds, we optimized the extrinsic parameters by minimize 3D-3D point-to-point error (i.e. $h_{pt}$). Therefore, the experiments are set to examine, whether the theory is feasible and effective in the scenario when point-to-point error (i.e. $h_{pt}$) is considered. In order to verify the conclusions of the theoretical derivation, we obtained two sets of data for seven calibration targets centralized and scattered in a certain field of view for polygonal method and box method.}}

\begin{figure}[tp]
     \centering
      \includegraphics[scale=0.32]{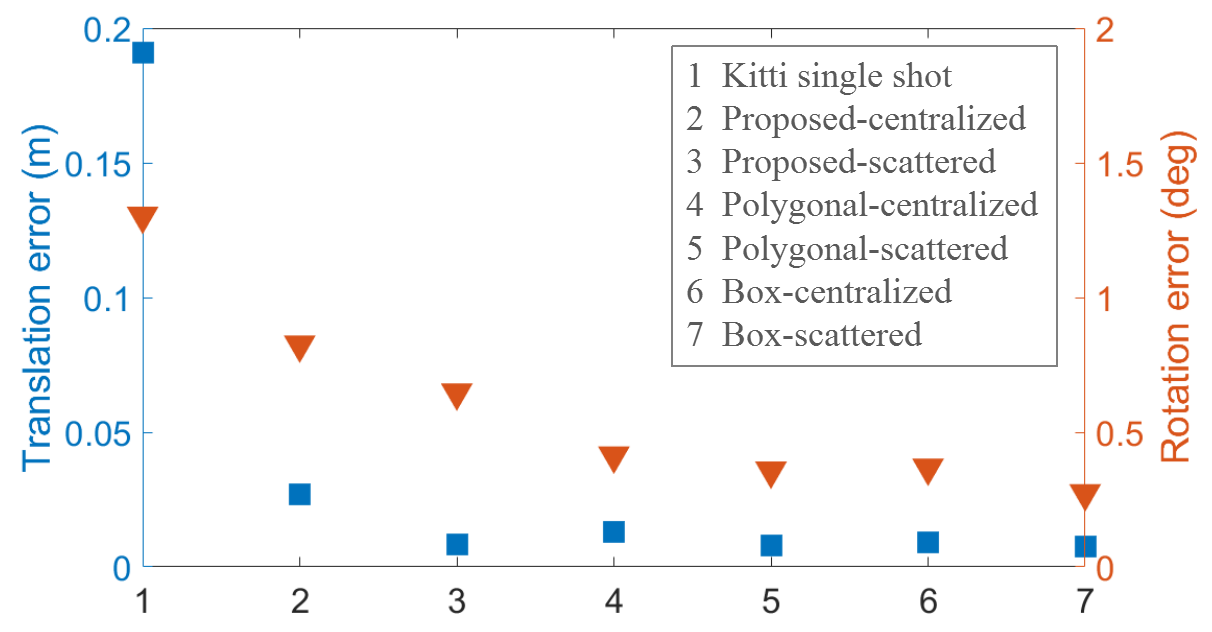}
      \caption{{\color{blue}{Accuracy comparison results: errors of the calibration result by single shot method, proposed method, polygonal method and box method.}}}
      \label{3D}
\end{figure}

In the theoretical derivation part, we explored the influence of the placement of the calibration targets on the calibration accuracy, and concluded that the scattered placement of the calibration targets is beneficial for improving the calibration accuracy. The results, depicted in Fig. \ref{3D}, indicate that proposed method achieves better calibration results than the KITTI single shot method, because the camera's field of view limits the number of laser lines hitting the chessboards and the number of observed chessboards in the single shot method. {\color{blue}{It can be seen that polygons and boxes are the better calibration targets compared to chessboards. Besides, as reflected by the experiments, for each calibration target, the calibration results of the scattered placement are better than the centralized placement, which is consistent with the conclusion of Section IV.}}
\subsection{Real-World Experiment}
We conducted real-world experiments and sufficiently compared the three different calibration methods: the KITTI single shot calibration method \cite{geiger2012automatic}; \textbf{MO methods}: multiple images and LiDAR data of a single calibration target presented in different directions as input which is fussy due to the need to move the chessboard, similar to \cite{zheng2015new}, \cite{zhang2017real}; and \textbf{the motion-based} calibration method: based on trajectory alignment.

\begin{figure}[tp]
     \centering
      \includegraphics[scale=0.36]{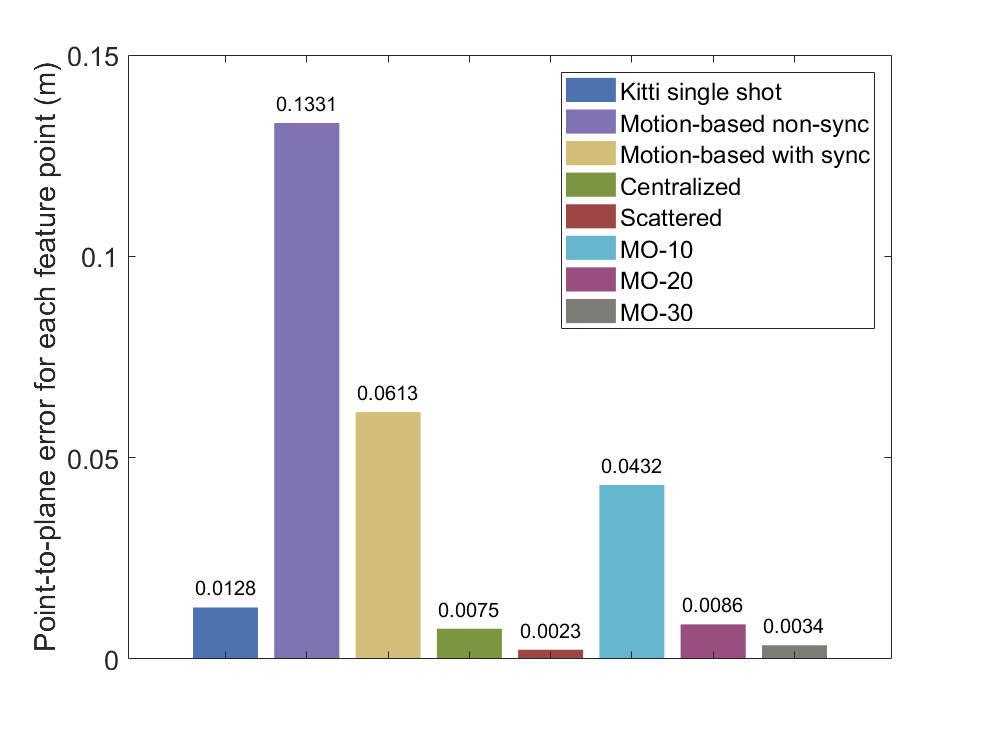}
      \caption{Real-world experiment results: calibration errors for single shot method, motion-based method, MO method and our method.}
      \label{calibrate}
\end{figure}

We obtained one set of data for the KITTI single shot, two sets of data in which the chessboards were placed scattered and centralized for our method, two sets of data for the motion-based calibration, and one set of data for the MO calibration method, including 67 corresponding laser scans and camera image data under different poses. The 67 pairs of laser scans and camera images were divided into two parts. The first 30 data pairs were used for the MO method to calibrate the extrinsic parameters, and the last 37 remaining data pairs were used to test the accuracy of all methods. As mentioned before, we evaluated the calibration results estimated by different methods with the point-to-plane error using the same data. The calibration error is shown in Fig. \ref{calibrate}.
We made two experiments for the motion-based method, one is called the \textbf{Motion-based non-sync method}, for which we did not complete the hardware synchronization between the two sensors. The other is called the \textbf{Motion-based with sync method}, for which we completed a rough hardware synchronization by finding the nearest neighbor on the timestamps of the two sensors. For the motion-based calibration method, time synchronization is not easy; even if a rough time synchronization is completed, the calibration result is not as good as with the appearance-based method.

The KITTI single shot calibration method does not give a very accurate result in the case of the Velodyne VLP-16 LiDAR. The calibration results of our scattered placement calibration method are better than the calibration results of the centralized placement, which is consistent with the conclusion of Section IV. The traditional appearance-based method is limited by the narrow common field of view, and the calibration accuracy is not as good as our method. This result proves once again that a larger common field of view between the two sensors obtained by our method can lead to a better calibration accuracy.

To discuss the calibration results of the MO calibration method, we did another set of experiments. The extrinsic calibration was performed using the data of 10 pairs (MO-10), 20 pairs (MO-20), and 30 pairs (MO-30), respectively. It can be seen that with more data used for calibration, the accuracy of the MO calibration method also increases and the MO calibration method achieves results that are only worse than our calibration method when using 30 angles. The MO method is time consuming to use and converges more slowly than our method. Since the MO method needs to have a common field of view between the sensors, this limits its use to some extent.

In summary, our method has the best calibration accuracy, and we only used the information of five chessboards. First, the extended field of view obtained by our method can improve the calibration accuracy, and our method can be applied to a case with an arbitrary configuration, which is beneficial and necessary in practical use.
{\color{blue}{The proposed method can be applied for calibrating multiple cameras and LiDAR devices. When proposed method is used to calibration between two LiDAR devices under arbitrary configurations, one of the LiDAR devices can be considered as a camera to run laser SLAM \cite{zhang2017low}. Second, our method eliminates the time variable from the spatial extrinsic parameters estimating, so it is applicable to the cases lacking time synchronization and will not introduce additional variables.}} That is to say, our error term does not include laser motion estimation error and time offset error. Third, as shown by our experiments, the calibration accuracy is higher in the case where the calibration targets are placed in a scattered manner, which is consistent with our theoretical derivation.

\begin{figure}[tp]
     \centering
      \includegraphics[scale=1.2]{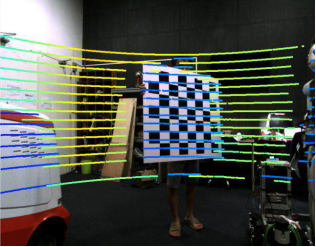}
      \caption{The results of our calibration of the laser data re-projected into the image. {\color{blue}{The yellow points on the chessboard are the points observed by LiDAR but not observed by the camera due to the occlusion.}}}
      \label{reprojected}
\end{figure}

Finally, we show the results of our method of the laser data reprojected into the image, which allows us to see the accuracy of the calibration more intuitively, as shown in Fig. \ref{reprojected}.

\section{Conclusion}
We proposed a LiDAR-camera extrinsic calibration method eliminated the time variable and the limitation of sharing a common field of view. Furthermore, we analyzed the observability of the calibration system and derived how the calibration targets can be placed better to improve the accuracy of the extrinsic calibration. Then we made a full comparison with other methods through both simulation and real-world experiments, which showed that our method can give a higher precision calibration result. In order to simplify the calibration, our next work is to study a calibration method that does not require a calibration target and still maintains calibration accuracy.
\appendix{
\subsection{Observability Of Our Calibration System With $h_{pl}$}
What follows is the analysis about the minimal necessary conditions of the chessboards setting to solve the accurate 6DoF extrinsic calibration problem.

\textbf{Observation of one plane:} Suppose there are enough points on each chessboard plane, the observability matrix of chessboard's feature $i$ and feature $j$ at time $t_{k}$ becomes:
\begin{equation}\label{eq.sub30}
  {M}_{(i,j)k}\triangleq
  \begin{bmatrix}
  \breve{J}_{ik}\hat{R}_{k}\begin{bmatrix} \breve{\Gamma}_{ik} & -I_{3} & 0_{3} & \ldots & I_{3} & \ldots & 0_{3} \end{bmatrix} \\
  \breve{J}_{jk}\hat{R}_{k}\begin{bmatrix} \breve{\Gamma}_{jk} & -I_{3} & 0_{3} & \ldots & \ldots & I_{3} & 0_{3} \end{bmatrix} \\
  \begin{bmatrix} 0 & 0 & 0 & \ldots & n_{r}^{T} & \ldots & 0   \end{bmatrix}\\
  \begin{bmatrix} 0 & 0 & 0 & \ldots & \ldots & n_{r}^{T} & 0   \end{bmatrix}
  \end{bmatrix}
\end{equation}

Then, we can obtain:
\begin{multline}\label{eq.sub31} {M}_{(i,j)k}\breve{N}=\left[\begin{array}{cccc}
  0_{4\times1}&0_{4\times1}&0_{4\times1}&0_{4\times1} \\
  n_{1}&n_{2}&n_{3}&n_{3}p_{i2}-n_{2}p_{i3} \\
  n_{1}&n_{2}&n_{3}&n_{3}p_{j2}-n_{2}p_{j3}
  \end{array}\right.\\
\left.\begin{array}{cc}
  0_{4\times1} & 0_{4\times1} \\
  -n_{3}p_{i1}+n_{1}p_{i3} & n_{2}p_{i1}-n_{1}p_{i2} \\
  -n_{3}p_{j1}+n_{1}p_{j3} & n_{2}p_{j1}-n_{1}p_{j2}
  \end{array}\right]
\end{multline}%
where $n_{r}^{T}\triangleq\begin{bmatrix} n_{1} & n_{2} & n_{3} \end{bmatrix}$, $^{\mathcal{L}}p_{f_{i}}\triangleq\begin{bmatrix} p_{i1} & p_{i2} & p_{i3} \end{bmatrix}$,$^{\mathcal{L}}p_{f_{j}}\triangleq\begin{bmatrix} p_{j1} & p_{j2} & p_{j3} \end{bmatrix}$. Next, we perform an elementary linear transformation on the above equation. That is, both sides of (\ref{eq.sub31}) are multiplied by matrix $A_{1}$.
\begin{equation}\label{eq.sub32}
  A_{1}\triangleq\begin{bmatrix} 1 & -\frac{n_{2}}{n_{1}} & -\frac{n_{3}}{n_{1}} & 0 & 0 & 0 \\
   0 & 1 & 0 & 0 & 0 & 0 \\
   0 & 0 & 1 & 0 & 0 & 0 \\
   0 & 0 & 0 & 1 & 0 & n_{1} \\
   0 & 0 & 0 & 0 & 1 & n_{2} \\
   0 & 0 & 0 & 0 & 0 & n_{3} \end{bmatrix}
\end{equation}
\begin{multline}\label{eq.sub33} {M}_{(i,j)k}\breve{N}A_{1}=\left[\begin{array}{cccc}
  0_{4\times1}&0_{4\times1}&0_{4\times1}&0_{4\times1} \\
  n_{1}&0&0&n_{3}p_{i2}-n_{2}p_{i3} \\
  n_{1}&0&0&n_{3}p_{j2}-n_{2}p_{j3}
  \end{array}\right.\\
\left.\begin{array}{cc}
  0_{4\times1} & 0_{4\times1} \\
  -n_{3}p_{i1}+n_{1}p_{i3} & 0 \\
  -n_{3}p_{j1}+n_{1}p_{j3} & 0
  \end{array}\right]
\end{multline}

As shown, there are three columns of ${M}_{(i,j)k}\breve{N}A_{1}$ that become all zeros. So the second, third and sixth columns of $\breve{N}A_{1}$ are the nullspace of $M_{k}$, we denote it as $N_{1}$ and can get:
\begin{equation}\label{eq.sub222222222222229}
  M_{(i,j)k}N_{1}=0_{6\times3}
\end{equation}

Since this holds for any $i$, $j$ and any $k$, we conclude that $MN_{1}=0$. Note that the first three columns of $\breve{N}A_{1}$ correspond to global translations of the state vector, while the last three columns to global rotations. Therefore, when observing only one plane, any translation parallel to the plane's normal and any rotation around the plane's normal vector is unobservable.

\textbf{Observation of two planes:} the observability matrix of feature $i$ and feature $j$ at time $t_{k}$ from two chessboards, described by $n_{r_{a}}$,and $n_{r_{b}}$:
\begin{equation}\label{eq.sub34}
  {M}_{(i,j)k}\triangleq
  \begin{bmatrix}
  \breve{J}_{ik}\hat{R}_{k}\begin{bmatrix} \breve{\Gamma}_{ik} & -I_{3} & 0_{3} & \ldots & I_{3} & \ldots & 0_{3} \end{bmatrix} \\
  \breve{J}_{jk}\hat{R}_{k}\begin{bmatrix} \breve{\Gamma}_{jk} & -I_{3} & 0_{3} & \ldots & \ldots & I_{3} & 0_{3} \end{bmatrix} \\
  \begin{bmatrix} 0 & 0 & 0 & \ldots & n_{r_{a}}^{T} & \ldots & 0   \end{bmatrix}\\
  \begin{bmatrix} 0 & 0 & 0 & \ldots & \ldots & n_{r_{b}}^{T} & 0   \end{bmatrix}
  \end{bmatrix}
\end{equation}

Then, we can obtain:
\begin{multline}\label{eq.sub35} {M}_{(i,j)k}\breve{N}=\left[\begin{array}{cccc}
  0_{4\times1}&0_{4\times1}&0_{4\times1}&0_{4\times1} \\
  n_{a_{1}}&n_{a_{2}}&n_{a_{3}}&n_{a_{3}}p_{i2}-n_{a_{2}}p_{i3} \\
  n_{b_{1}}&n_{b_{2}}&n_{b_{3}}&n_{b_{3}}p_{j2}-n_{b_{2}}p_{j3}
  \end{array}\right.\\
\left.\begin{array}{cc}
  0_{4\times1} & 0_{4\times1} \\
  -n_{a_{3}}p_{i1}+n_{a_{1}}p_{i3} & n_{a_{2}}p_{i1}-n_{a_{1}}p_{i2} \\
  -n_{b_{3}}p_{j1}+n_{b_{1}}p_{j3} & n_{b_{2}}p_{j1}-n_{b_{1}}p_{j2}
  \end{array}\right]
\end{multline}
where $n_{r_{a}}^{T}\triangleq\begin{bmatrix} n_{a_{1}} & n_{a_{2}} & n_{a_{3}} \end{bmatrix}$,$n_{r_{b}}^{T}\triangleq\begin{bmatrix} n_{b_{1}} & n_{b_{2}} & n_{b_{3}} \end{bmatrix}$.

Next, we perform an elementary linear transformation on the above equation. That is, both sides of (\ref{eq.sub35}) are multiplied by matrix $A_{2}$.
\begin{equation}\label{eq.sub36}
  A_{2}\triangleq\begin{bmatrix} 1 & -\frac{n_{a_{2}}}{n_{a_{1}}} & \frac{n_{a2}}{n_{a1}}\cdot \Lambda-\frac{n_{a3}}{n_{a1}} & 0 & 0 & 0 \\
   0 & 1 & -\Lambda & 0 & 0 & 0 \\
   0 & 0 & 1 & 0 & 0 & 0 \\
   0 & 0 & 0 & 1 & 0 & 0 \\
   0 & 0 & 0 & 0 & 1 & 0 \\
   0 & 0 & 0 & 0 & 0 & 1 \end{bmatrix}
\end{equation}
\begin{equation}\label{eq.sub37}
  \Lambda\triangleq\frac{n_{b3}n_{a1}-n_{b1}n_{a3}}{n_{b2}n_{a1}-n_{b1}n_{a2}}
\end{equation}
\begin{multline}\label{eq.sub38} {M}_{(i,j)k}\breve{N}A_{2}=\left[\begin{array}{ccc}
  0_{4\times1}&0_{4\times1}&0_{4\times1} \\
  n_{a_{1}}&0&0 \\
  n_{b_{1}}& n_{b2}-\frac{n_{b1}n_{a2}}{n_{a1}} &0
  \end{array}\right.\\
\left.\begin{array}{ccc}
  0_{4\times1}&0_{4\times1} & 0_{4\times1} \\
  n_{a_{3}}p_{i2}-n_{a_{2}}p_{i3}&-n_{a_{3}}p_{i1}+n_{a_{1}}p_{i3} & n_{a_{2}}p_{i1}-n_{a_{1}}p_{i2} \\
  n_{b_{3}}p_{j2}-n_{b_{2}}p_{j3}&-n_{b_{3}}p_{j1}+n_{b_{1}}p_{j3} & n_{b_{2}}p_{j1}-n_{b_{1}}p_{j2}
  \end{array}\right]
\end{multline}

As shown, the third column of the $\breve{N}A_{2}$ is the nullspace of $M_{(i,j)k}$, we denote it as $N_{2}$ and can get:
\begin{equation}\label{eq.sub2222222222222229}
  M_{(i,j)k}N_{2}=0_{6\times1}
\end{equation}

Therefore, when observing two planes, one degree of freedom of the translation is unobservable.

\textbf{Observation of three planes:} Similar to the previous derivation process, we conclude that when three planes with non-collinear normal vectors are observed, we can determine all the unknowns. That is to say, our calibration system is observable.

\subsection{Observability Of Our Calibration System With $h_{pt}$}
{\color{blue}{Our calibration method can use different calibration targets, for example, polygonal planar boards \cite{park2014calibration} or boxes \cite{pusztai2017accurate}. For the data obtained by above calibration targets, the error measurement is the point-to-point error measurement $h_{pt}$:}}
\begin{equation}\label{eq.sub2288}
  h_{pt}={^{\mathcal{L}}p_{c}}-p_{r}
\end{equation}

{\color{blue}{Thus, the measurement Jacobian matrix $H_{ik}$ at time $t_{k}$ for feature $i$ is given by:}}
\begin{equation}\label{eq.sub128}
  {H}_{ik}\triangleq\begin{bmatrix} \breve{H}_{ik} \\ H_{pt} \end{bmatrix}=\begin{bmatrix} \breve{H}_{I_{ik}} & 0 & \ldots & \breve{H}_{f_{ik}} & \ldots & 0_{3\times1} \\ 0_{3\times1} & 0_{3\times1} & \ldots & I_{3} & \ldots & 0_{3\times1}\end{bmatrix}
\end{equation}
\begin{equation}\label{eq.sub22222228}
  H_{pt}{\color{blue}{\triangleq \frac{\partial h_{pt}}{\partial x_{k}} }}
\end{equation}
{\color{blue}{where $H_{pt}$ refers to the Jacobian matrix of $h_{pt}$ with respect to $^{\mathcal{L}}x_{c}$ and ${^{\mathcal{L}}p_{f}}$ (the point belonging to the chessboard is represented as $^{\mathcal{L}}p_{c}$). What follows is the analysis about the minimal necessary conditions of the point pairs setting to solve the accurate 6DoF extrinsic calibration problem.}}

\textbf{Observation of one point:} {\color{blue}{Suppose there are enough points on each calibration target, the observability matrix of calibration target's feature $i$ at time $t_{k}$ becomes:}}
\begin{equation}\label{eq.sub130}
  {M}_{ik}\triangleq
  \begin{bmatrix}
  \breve{J}_{ik}\hat{R}_{k}\begin{bmatrix} \breve{\Gamma}_{ik} & -I_{3} & 0_{3} & \ldots & I_{3} & \ldots & 0_{3} \end{bmatrix} \\
  \begin{bmatrix} 0_{3\times1} & 0_{3\times1} & 0_{3\times1} & \ldots & I_{3} & \ldots & 0_{3\times1}   \end{bmatrix}
  \end{bmatrix}
\end{equation}

{\color{blue}{Then, we can obtain:}}
\begin{equation}\label{eq.sub131}
{M}_{ik}\breve{N}=
\begin{bmatrix}
0_{2\times1}&0_{2\times1}&0_{2\times1}&0_{2\times1} &0_{2\times1} & 0_{2\times1} \\
  1&0&0&0 &p_{i3} & -p_{i2} \\
  0&1&0&-p_{i3}&0 & p_{i1} \\
  0&0&1&p_{i2} &-p_{i1} & 0
  \end{bmatrix}
\end{equation}
{\color{blue}{where $^{\mathcal{L}}p_{f_{i}}\triangleq\begin{bmatrix} p_{i1} & p_{i2} & p_{i3} \end{bmatrix}$. Next, we perform an elementary linear transformation on the above equation. That is, both sides of (\ref{eq.sub131}) are multiplied by matrix $A_{3}$.}}
\begin{equation}\label{eq.sub132}
  A_{3}\triangleq\begin{bmatrix}
   1 & 0 & 0 & 0 & -p_{i3} & p_{i2} \\
   0 & 1 & 0 & p_{i3} & 0 & -p_{i1} \\
   0 & 0 & 1 & -p_{i2} & p_{i1} & 0 \\
   0 & 0 & 0 & 1 & 0 & 0 \\
   0 & 0 & 0 & 0 & 1 & 0 \\
   0 & 0 & 0 & 0 & 0 & 1 \end{bmatrix}
\end{equation}
\begin{equation}\label{eq.sub133}
{M}_{ik}\breve{N}A_{3}=
\begin{bmatrix}
0_{2\times1}&0_{2\times1}&0_{2\times1}&0_{2\times1} &0_{2\times1} & 0_{2\times1} \\
  1&0&0&0 &0 & 0 \\
  0&1&0&0&0 & 0 \\
  0&0&1&0 &0 & 0
  \end{bmatrix}
\end{equation}

{\color{blue}{As shown, there are three columns of ${M}_{ik}\breve{N}A_{3}$ that become all zeros. So the fourth, fifth and sixth columns of $\breve{N}A_{3}$ are the nullspace of $M_{k}$, we denote it as $N_{3}$ and can get:
\begin{equation}\label{eq.sub2222222222222229}
  M_{ik}N_{3}=0_{5\times3}
\end{equation}

Since this holds for any $i$, $j$ and any $k$, we conclude that $MN_{3}=0$. Therefore, when observing only one point, any rotation is unobservable.}}

\textbf{Observation of two points:} {\color{blue}{the observability matrix of feature $i$ and feature $j$ at time $t_{k}$:}}
\begin{equation}\label{eq.sub134}
  {M}_{(i,j)k}\triangleq
  \begin{bmatrix}
  \breve{J}_{ik}\hat{R}_{k}\begin{bmatrix} \breve{\Gamma}_{ik} & -I_{3} & 0_{3} & \ldots & I_{3} & \ldots & 0_{3} \end{bmatrix} \\
  \breve{J}_{jk}\hat{R}_{k}\begin{bmatrix} \breve{\Gamma}_{jk} & -I_{3} & 0_{3} & \ldots & \ldots & I_{3} & 0_{3} \end{bmatrix} \\
  \begin{bmatrix} 0_{3\times1} & 0_{3\times1} & 0_{3\times1} & \ldots & I_{3} & \ldots & 0_{3\times1}   \end{bmatrix}\\
  \begin{bmatrix} 0_{3\times1} & 0_{3\times1} & 0_{3\times1} &  \ldots & \ldots & I_{3} & 0_{3\times1}   \end{bmatrix}
  \end{bmatrix}
\end{equation}

{\color{blue}{Then, we can obtain:}}
\begin{equation}\label{eq.sub135}
{M}_{(i,j)k}\breve{N}=
\begin{bmatrix}
0_{4\times1}&0_{4\times1}&0_{4\times1}&0_{4\times1} &0_{4\times1} & 0_{4\times1} \\
  1&0&0&0 &p_{i3} & -p_{i2} \\
  0&1&0&-p_{i3}&0 & p_{i1} \\
  0&0&1&p_{i2} &-p_{i1} & 0 \\
  1&0&0&0 &p_{j3} & -p_{j2} \\
  0&1&0&-p_{j3}&0 & p_{j1} \\
  0&0&1&p_{j2} &-p_{j1} & 0
  \end{bmatrix}
\end{equation}
{\color{blue}{where $^{\mathcal{L}}p_{f_{j}}\triangleq\begin{bmatrix} p_{j1} & p_{j2} & p_{j3} \end{bmatrix}$. Next, we perform an elementary linear transformation on the above equation. That is, both sides of (\ref{eq.sub135}) are multiplied by matrix $A_{4}$.}}
\begin{equation}\label{eq.sub136}
  A_{4}\triangleq\begin{bmatrix}
   1 & 0 & 0 & \frac{-p_{i2}p_{j3}+p_{i3}p_{j2}}{p_{i1}-p_{j1}} & -p_{j3} & p_{j2} \\
   0 & 1 & 0 & p_{j3}-\frac{p_{j1}(p_{i2}-p_{j2})}{p_{i1}-p_{j1}} & 0 & -p_{j1} \\
   0 & 0 & 1 & -p_{j2}+\frac{p_{j1}(p_{i3}-p_{j3})}{p_{i1}-p_{j1}} & p_{j1} & 0 \\
   0 & 0 & 0 & 1 & 0 & 0 \\
   0 & 0 & 0 & \frac{p_{i2}-p_{j2}}{p_{i1}-p_{j1}} & 1 & 0 \\
   0 & 0 & 0 & \frac{p_{i3}-p_{j3}}{p_{i1}-p_{j1}} & 0 & 1 \end{bmatrix}
\end{equation}


\begin{multline}\label{eq.sub138} {M}_{(i,j)k}\breve{N}A_{3}=\left[\begin{array}{ccc}
  0_{4\times1}&0_{4\times1}&0_{4\times1} \\
  1&0&0 \\
  0&1&0 \\
  0&0&1 \\
  1&0&0 \\
  0&1&0 \\
  0&0&1
  \end{array}\right.\\
\left.\begin{array}{ccc}
  0_{4\times1}&0_{4\times1} & 0_{4\times1} \\
  0 &p_{i3}-p_{j3} & -p_{i2}+p_{j2} \\
  0&0 & p_{i1}-p_{j1}\\
  0 &-p_{i1}+p_{j1} & 0 \\
  0 &0 & 0 \\
  0 &0 & 0\\
  0 &0 & 0 \\
  \end{array}\right]
\end{multline}
{\color{blue}{As shown, the fourth column of the $\breve{N}A_{4}$ is the nullspace of $M_{(i,j)k}$, we denote it as $N_{4}$ and can get:
\begin{equation}\label{eq.sub2222222222222229}
 M_{(i,j)k}N_{4}=0_{10\times1}
\end{equation}

Therefore, when observing two points, one degree of freedom of the rotation is unobservable.}}

\textbf{Observation of three points:} {\color{blue}{Similar to the previous derivation process, we conclude that when three non-collinear points are observed, we can determine all the unknowns.}}
}


%

%
%
%
%
%

\ifCLASSOPTIONcaptionsoff
  \newpage
\fi

\bibliographystyle{ieeetr} 
\bibliography{fubo}




\begin{IEEEbiography}[{\includegraphics[width=0.90in,clip,keepaspectratio]{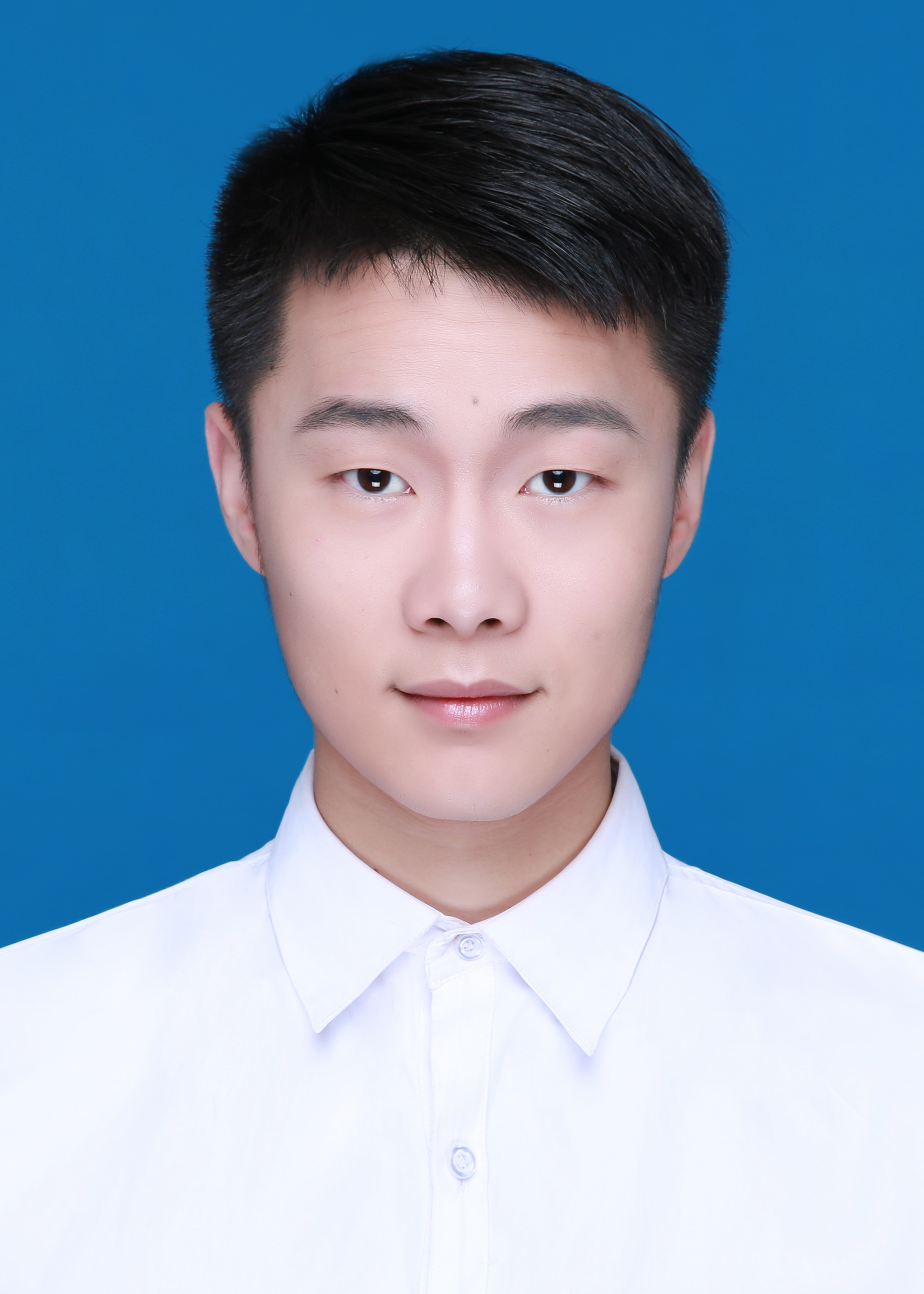}}]{Bo Fu} received his BS from the Department of Control Science and Engineering, Shandong University, Jinan, P.R. China in 2017. \\
\quad He is currently an MS candidate in the Department of Control Science and Engineering, Zhejiang University, Hangzhou, P.R. China. His latest research interests include multisensor calibration and sensor fusion.
\end{IEEEbiography}
\vspace{-0.01cm}
\begin{IEEEbiography}[{\includegraphics[width=0.90in,clip,keepaspectratio]{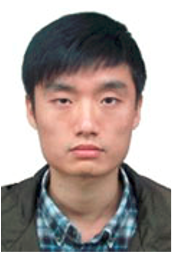}}]{Yue Wang} received his PhD the from Department of Control Science and Engineering, Zhejiang University, Hangzhou, P.R. China in 2016. \\
\quad He is currently a lecturer in the Department of Control Science and Engineering, Zhejiang University, Hangzhou, P.R. China. His latest research interests include mobile robotics and robot perception.
\end{IEEEbiography}
\vspace{-0.01cm}
\begin{IEEEbiography}[{\includegraphics[width=0.90in,clip,keepaspectratio]{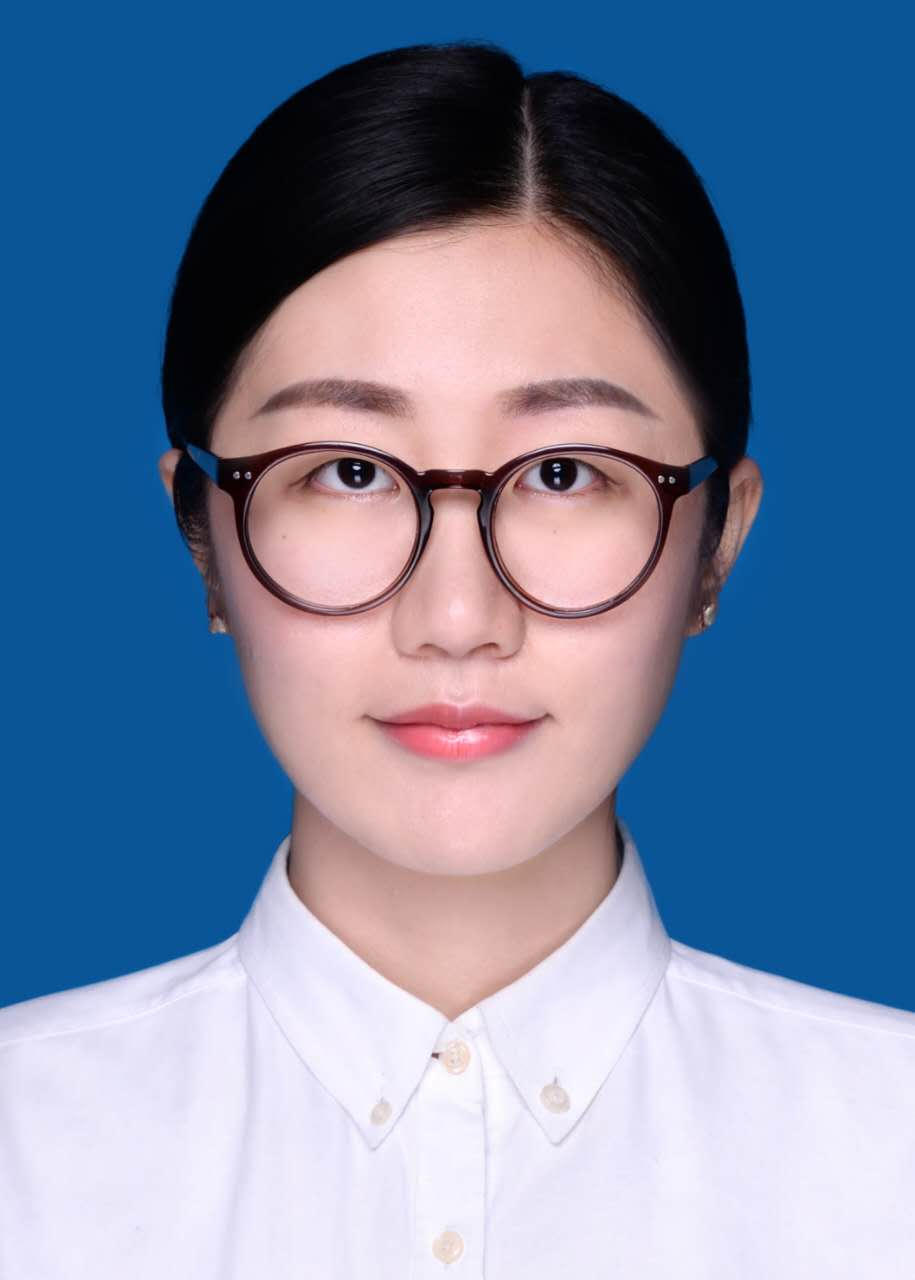}}]{Xiaqing Ding} received her BS from the Department of Control Science and Engineering, Zhejiang University, Hangzhou, P.R. China in 2016. \\
\quad She is currently a Ph.D. candidate in the Department of Control Science and Engineering, Zhejiang University, Hangzhou, P.R. China. Her latest research interests include SLAM and vision based localization.
\end{IEEEbiography}
\vspace{-0.01cm}
\begin{IEEEbiography}[{\includegraphics[width=0.90in,clip,keepaspectratio]{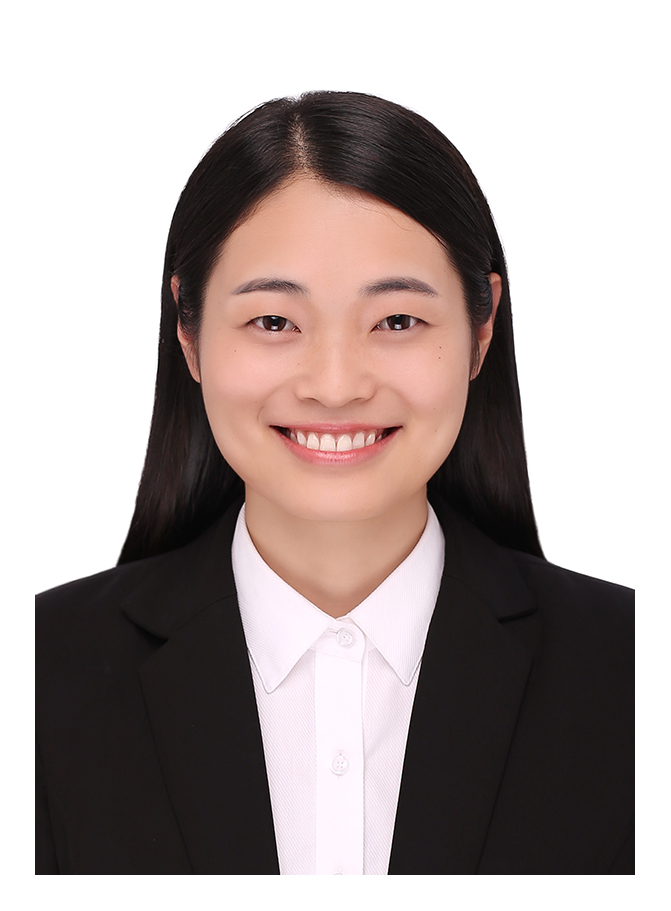}}]{Yanmei Jiao} received her BS from the Department of Computer Science and Control Engineering, Nankai University, Tianjin, P.R. China in 2017. \\
\quad She is currently a Ph.D. candidate in the Department of Control Sciencec and Engineering, Zhejiang University, Hangzhou, P.R. China. Her research interests include computer vision and vision based localization.
\end{IEEEbiography}
\vspace{-0.01cm}
\begin{IEEEbiography}[{\includegraphics[width=0.90in,clip,keepaspectratio]{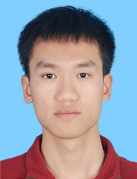}}]{Li Tang} received his BS from the Department of Control Science and Engineering, Zhejiang University, Hangzhou, P.R. China in 2015. \\
\quad He is currently a Ph.D. candidate in the Department of Control Science and Engineering, Zhejiang University, Hangzhou, P.R. China. His research interests include vision based localization and autonomous navigation.
\end{IEEEbiography}
\vspace{-0.01cm}
\begin{IEEEbiography}[{\includegraphics[width=0.90in,clip,keepaspectratio]{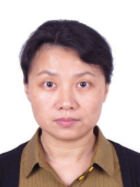}}]{Rong Xiong} received her PhD from the Department of Control Science and Engineering, Zhejiang University, Hangzhou, P.R. China in 2009. \\
\quad She is currently a professor in the Department of Control Science and Engineering, Zhejiang University, Hangzhou, P.R. China. Her latest research interests include motion planning and SLAM.
\end{IEEEbiography}

\end{document}